\documentclass[a4paper,fleqn]{cas-sc}

\usepackage[authoryear,longnamesfirst]{natbib}

\usepackage{arydshln}
\usepackage{amsmath}
\usepackage{graphicx}
\usepackage[bb=boondox]{mathalfa}
\DeclareMathOperator*{\argmax}{arg\,max}
\DeclareMathOperator*{\argmin}{arg\,min}

\def\tsc#1{\csdef{#1}{\textsc{\lowercase{#1}}\xspace}}
\tsc{WGM}
\tsc{QE}
\tsc{EP}
\tsc{PMS}
\tsc{BEC}
\tsc{DE}

\begin{document}
\let\WriteBookmarks\relax
\def\floatpagepagefraction{1}
\def\textpagefraction{.001}

\shorttitle{$DefSent+$}

\shortauthors{Xiaodong Liu}

\title [mode = title]{DefSent\LARGE\texttt{+}: Improving sentence embeddings of language models by projecting definition sentences into a quasi-isotropic or isotropic vector space of unlimited dictionary entries}                      



%
\author[1]{Xiaodong Liu}[orcid=0009-0004-6648-2302]

\cormark[1]


\ead{xiaodongliu@ist.hokudai.ac.jp}


\credit{Conceptualization, Data curation, Methodology, Resources, Software, Investigation, Validation, and Writing - Original Draft}

\affiliation[1]{organization={Language Media Lab, Graduate School of Information Science and Technology, Hokkaido University},
            addressline={Kita 8, Nishi 5, Kita-ku}, 
            city={Sapporo},
            postcode={060-0808}, 
            state={Hokkaido},
            country={Japan}}

\cortext[cor1]{Corresponding author}



\begin{abstract}
This paper presents a significant improvement on the previous conference paper known as $DefSent$. The prior study seeks to improve sentence embeddings of language models by projecting definition sentences into the vector space of dictionary entries. We discover that this approach is not fully explored due to the methodological limitation of using word embeddings of language models to represent dictionary entries. This leads to two hindrances. First, dictionary entries are constrained by the single-word vocabulary, and thus cannot be fully exploited. Second, semantic representations of language models are known to be anisotropic, but pre-processing word embeddings for $DefSent$ is not allowed because its weight is frozen during training and tied to the prediction layer. In this paper, we propose a novel method to progressively build entry embeddings not subject to the limitations. As a result, definition sentences can be projected into a quasi-isotropic or isotropic vector space of unlimited dictionary entries, so that sentence embeddings of noticeably better quality are attainable. We abbreviate our approach as $DefSent+$ (a plus version of $DefSent$), involving the following strengths: 1) the task performance on measuring sentence similarities is significantly improved compared to $DefSent$; 2) when $DefSent+$ is used to further train data-augmented models like SIMCSE, SNCSE, and SynCSE, state-of-the-art performance on measuring sentence similarities can be achieved among the approaches without using manually labeled datasets; 3) $DefSent+$ is also competitive in feature-based transfer for NLP downstream tasks.
\end{abstract}




\maketitle

\section{Introduction}
\label{sec:introduction}
Sentence embeddings are dense vector representations that convey semantic information for large chunks of text such as sentences. To begin with, it should be clarified that we follow two conventions in the research field of sentence embeddings. In this paper, sentence embeddings are represented by pooling strategies \citep{sbert} applied to token representations output from auto-encoding language models such as BERT \citep{bert} and RoBERTa \citep{roberta}. Besides, the major application of our sentence embeddings is same as SBERT \citep{sbert}: to store a large number of sentences as vector representations in memory for searching most semantically related sentence pairs in a fast manner.

\subsection{Background}
\label{sec:background}
As token representations output from raw pre-trained BERT and RoBERTa suffer from anisotropic vector spaces \citep{anisotropicwordembeddingsregularization,selfsim}, sentence embeddings encoded by raw pre-trained models via pooling strategies tend to carry similar semantic features for all sentences, and consequently poorly distinguish semantic similarities of sentence pairs, which is experimentally shown in \citet{sbert}'s work. To improve the sentence embeddings, there are three categories of fine-tuning raw pre-trained models:
\begin{enumerate}
    \item supervised approaches using large labeled Natural Language Inference (NLI) datasets like SNLI \citep{snli} to fine-tune raw pre-trained models by siamese architectures \citep{sbert,simcse};

    \item unsupervised learning such as mapping sentence embeddings to an isotropic space \citep{bert-flow,whitening} or data augmentation using contrastive learning \citep{simcse,esimcse,sncse}\footnote{It can be argued that this approach is contrastive self-supervised learning by seeking methodology to generate positive or negative sentence pairs without manual labeling. But we follow the definition --- unsupervised contrastive learning claimed in their papers. No matter which stands more valid, there is no doubt that manually labeled datasets are not used.};

    \item self-supervised learning by projecting definition sentences into the vector spaces of dictionary entries \citep{defsent} (supervisory signals are entry ids, which do not require manual labeling).
\end{enumerate}
Among the approaches not replying on large manually labeled datasets, we find that the potential of \citet{defsent}'s approach lies buried and is promising. Generally, this paper seeks to excavate the full potential of this approach, so that sentence embeddings of noticeably better quality are attainable. We introduce the details in the following subsections.

\subsection{Mechanism, motivation \& research Goal}
\label{sec:motivation}
Given a dictionary dataset containing entries $e^{'}$, for any entry-sentence pair ($e$, $s$) in the dataset, modeling $logp_{\theta}(e|s)$ is formulated by Equation \ref{equ:softmax}, where $h_{\theta}(s)$ is the representation output from a model to approximate $entry\_embed(e)$ and $\theta$ is trainable model parameters. $h_{\theta}(s)$ can also be interpreted as the representation projected into the vector space of dictionary entries.
\begin{equation}
     \argmax\limits_{\theta}logp_{\theta}(e|s) = \argmin\limits_{\theta} -log \frac{exp(entry\_embed(e) \cdot h_{\theta}(s))} {\sum\limits_{e^{'}} exp(entry\_embed(e^{'}) \cdot h_{\theta}(s))}
\label{equ:softmax}
\end{equation}
It should be underlined that entry embeddings are \underline{\textbf{frozen}} during training, and the embeddings have their own vector space. Thus, under the same optimizer setting\footnote{In this paper, we follow the default setting \citep{bert} of fine-tuning pre-trained models.}, how well $\theta$ is optimized hinges on what sort of vector space that $h_{\theta}(s)$ will be projected into. \emph{For this consideration, if seeking to improve sentence embeddings by this approach, we should first shed focus on the quality of entry embeddings}. Let $R^{n\times d}$ denote the vector space of entry embeddings, where $n$ is the number of entries and $d$ is the number of dimensionalities $\mathbb{d}$ (i.e., the semantic axes of random variables such as ``food'', ``machine'', etc.). These two contributing factors, $n$ and $\mathbb{d}$, determine the quality of entry embeddings. With a comparatively big $n$, it implies that more entry-sentence pairs can be used for training, suggesting that more sentences can be projected into the $R$ for learning their semantic relatedness. This is conducive to generalization for unseen sentences. Also, a big $n$ indicates dense entry distributions on each $\mathbb{d}$, approaching the $R$ of a real-word scenario. It is helpful for projected sentences to learn more accurate semantic relatedness. As for $\mathbb{d}$, it is proved by the previous works \citep{bert-flow,whitening} that mapping sentence embeddings to an isotropic space, namely the whitened space where distributions on semantic axes are decorelated and have unit variance, can improve their quality. A recent work \citep{ICA-transform} further shows that remaining anisotropic information after whitening can be addressed if distributions on semantic axes are independent\footnote{In statistical sense, ``independent'' is stronger than ``decorrelated'', as non-linear correlation is also addressed if any two random variables are independent. For an example, see \citet{ICA-non-linear-correlation}.} from each other.

In the existing implementation of this approach, $DefSent$ \citep{defsent} utilizes the word embeddings of language models as entry embeddings. This choice entails the limitations explained in the following two paragraphs.

\textbf{Limited $\boldsymbol{n}$.} If the word embeddings are utilized as entry embeddings, dictionary entries are inevitably constrained by the single-word vocabulary\footnote{Single-word vocabulary indicates complete words occupying their own token ids in the word embeddings of language models. Two opposites are words comprising sub-words (e.g., ``jubilant'' composed of ``ju'', ``\#\#bil'', and ``\#\#ant'' in BERT vocabulary) and phrases comprising multiple single words or sub-words.} of language models. For instance, \citet{defsent} use the Oxford dictionary resource containing 36,767 dictionary entries, among which, only 9,647 entries are covered by BERT vocabulary as single words. The exploitation rate of the dictionary resource is 26.24\% for their training architecture; in other words, $DefSent$ is unable to predict the rest 73.76\% entries in this case. The rate will significantly decrease if multiple dictionary resources are combined together containing substantially more entries.

\textbf{Correlated $\boldsymbol{\mathbb{d}}$.} Semantic representations of language models are known to be anisotropic \citep{anisotropicwordembeddingsregularization,selfsim}. An indication of this is that the distribution of words is not directionally uniform in the vector space. This could result in word distributions on semantic axes having strong covariances, suggesting either positive or negative correlations. Consequently, there are not enough distinct semantic axes that can be combined to convey diverse semantics. A contrasting example can be observed in \citet{ICA-transform}'s paper where the distributions of words on axes are independent. In $DefSent$, pre-processing (e.g., whitening) word embeddings is not allowed, because the weight of word embeddings is tied to the decoder in the MLM head, and it is frozen during training. Once either of them is pre-processed, the modified weight of word embeddings will be unaligned to other model parameters in pre-trained models. This can make training difficult to converge.

For the considerations above, our research goal is to create entry embeddings not subject to the limitations, in order to observe whether sentence embeddings of better quality are attainable via this approach. Meanwhile, we also hold a research question worthy of investigation. Since sentence embeddings encoded by $DefSent$ encoders are quite competitive in the capability of feature-based transfer, it can be wondered that is the vector space, where entry distributions on different semantic axes are somewhat correlated, not completely negative for this approach? A whole investigation will be provided in Sections \ref{sec:sts} and \ref{sec:senteval} based on our experimental results, and we will conclude our findings in Section \ref{sec:conclusion}.

\subsection{Technical novelty}
\label{sec:novelty}
Given a dictionary dataset with $n$ entries, it is a small novelty that we manage to assign vector representations to all $n$ entries. The challenge is what can be done to make the vector space more directionally uniform than the anisotropic one. Let $X \in R^{n\times d}$ be the weight matrix of our entry embeddings. It is technically feasible to follow \citet{ICA-transform} to directly employ the algorithm of Independent Component Analysis (ICA) to pre-process the $X$, aiming to whiten the vector space and to further maximize the non-Gaussianity of distributions on semantic axes $\mathbb{d}$. However, we find that the direct use of ICA on the $X$ is not very effective for our training objective. In this paper, we propose a novel method to progressively build the $X$, with an anisotrpic vector space initially, but gradually evolve into a quasi-isotropic space. With this space as base, there are two types of vector space for definition sentences to project into: 
\begin{itemize}
    \item \textbf{quasi-isotropic}, entry distribution resembling a (rotated) ellipse in 2-d vector space, but not stringently directionally uniform;

    \item \textbf{isotropic (ICA-transformed)}, to whiten the quasi-isotropic vector space first, and then to maximize its non-Gaussianity.
\end{itemize}
Two types of vector space are effective for training different encoders. We present the relevant details in Section \ref{sec:method} describing our methodology.    

\subsection{Contributions}
\label{sec:contribution}
We abbreviate our approach as $DefSent+$ (a plus version of $DefSent$). The contributions of this paper lie in three folds, two with respect to the major application of sentence embeddings stated at the beginning of Section \ref{sec:introduction}, and one with respect to the capability of feature-based transfer.
\begin{enumerate}
    \item We offer a keen insight into the prior work known as $DefSent$ \citep{defsent}, and seek our methodology to address its limitations to fully explore its potentiality. In Section \ref{sec:sts}, we show that the task performance on Semantic Textual Similarity (STS) tasks are significantly improved compared to DefSent.

    \item Recent studies \citep{rankencoder} explore approaches to further training data-augmented models like SIMCSE \citep{simcse}, SNCSE \citep{sncse} and SynCSE \citep{syncse}. We also show that when $DefSent+$ is used for further training, state-of-the-art performance can be achieved on STS tasks, among the approaches without using manually labeled datasets.

    \item Among the previous approaches without using manually labeled datasets, $DefSent$ is comparatively weak on STS tasks, but shows stronger capability of feature-based transfer for downstream tasks. On the other hand, data-augmented models are relatively powerful on STS tasks, but not dominant on feature-based transfer. $DefSent+$ is competitive in both types of tasks. This follows the definition of good-quality sentence embeddings as suggested by the recent survey study \citep{SentenceEmbeddingSurvey}. 
\end{enumerate}

\section{Related work}
\label{sec:relatedwork}
In this section, we first concisely review the previous works about sentence embeddings. Then, we discuss the difference between our work and some related works in addition to $DefSent$ \citep{defsent}.

At the early stage, the vector representations for sentences can be simply realized by bag-of-words of word embeddings like GloVe \citep{glove} or count-based vector space models \citep{vsm}. Their performance on sentence similarity tasks is moderate nowadays. Neural network approaches can trace back to Skip-Thought \citep{skipthought}, an encoder-decoder architecture to reconstruct the surrounding sentences of an encoded passage. Later on, neural approaches start to rely on large labeled datasets like SNLI \citep{snli} for supervised learning. InferSent \citep{infersent} trains a siamese BiLSTM network with max-pooling followed by fully-connected layers to predict labels of sentence pairs. Universal Sentence Encoder \citep{universal} trains a transformer network \citep{transformer} and augments unsupervised learning with training on SNLI. Similar to \citet{universal}'s work, \citet{dan} proposes siamese DAN and siamese transformer networks trained on Reddit conversations, reinforced by multi-task learning including the use of SNLI. \citet{sbert} fine-tunes pre-trained language models \citep{bert,roberta} using a siamese architecture for labeled sentence pairs. Sentence embeddings output from SBERT or SRoBERTa can be used to search similar sentence pairs in a fast manner \citep{sbert}.

\citet{anisotropicwordembeddingsregularization} show that language models trained with tied input/output embedding weight are inclined to generate anisotropic word embeddings occupying a narrow cone in the vector space. \citet{selfsim} also finds that the problem exists in contextualized embeddings. Some works \citep{bert-flow,promptbert} point out that the inferiority is related to word frequency, subwords and case-sensitive. As a consequence, sentence embeddings encoded by raw pre-trained models result in inferior performance on semantic similarity tasks \citep{sbert}. To alleviate the problem, post-processing methods are proposed such as using regularization \citep{anisotropicwordembeddingsregularization,regularization2} and mapping to an isotropic space \citep{bert-flow,whitening}. Later, it is theoretically and empirically proved by \citet{simcse} that contrastive learning objective adopted by data augmentation methods can mitigate the problem effectively, which is a popular approach to improving the quality of sentence embeddings. The main idea of this approach is to generate positive pairs for given sentences while treat other independent sentences as negative pairs. With positive and negative pairs, InfoNCE \citep{infonce} loss is utilized to pull the embeddings of positive pairs closer, and push the embeddings of negative pairs farther. State-of-the-art models have devised interesting methods to sample positive pairs. \citet{simcse} pass any sentence to their encoder twice and perform dropout masking on their embeddings. \citet{esimcse} take into account sentence length for the drop masking. \citet{sncse} consider that concentrating solely on how to sample positive pairs can bring about feature suppression --- trained encoders cannot distinguish semantic similarity from textual similarity. To address the problem, \citet{sncse} incorporate soft negative pairs (explicit negation) into their training objective with the prompt technique \citep{promptbert} for sentence representations. The most recent RankEncoder \citep{rankencoder} leverages global context of Wikipedia corpus to generate rank vectors for sentence pairs, which can be used to further process other encoders.

Our work employs the training architecture of reverse dictionary to improve the sentence embeddings of language models, and the difference to $DefSent$ is described in Section \ref{sec:introduction}. Besides, the difference to some related works in terms of similar conceptualization should be discussed as well. First, as far as addressing anisotrpic word embeddings (entry embeddings in our case) are concerned, \citet{anisotropicwordembeddingsregularization} seek a theory that can explain the root cause during the pre-training stage of language models. They claim that during pre-training, most words are pulled towards a local region of the vector space while low-frequency words are pushed against the region, and thus a regularization method is proposed for pre-training to avoid the consequence. Our work, similar to \citet{ICA-transform}, is focused on post-processing separate static embeddings. Second, in regard to projecting sentences into isotropic vector space, \citet{whitening} apply their post-possessing procedure directly to sentences of target datasets. Our work, on the other hand, post-processes our entry embeddings first, which are then utilized as the media for projecting definition sentences. Our approach shows a better generalization for unseen sentences.

\section{Methodology}
\label{sec:method}
The method proposed by us aims to accomplish 1) full exploitation of dictionary resources, and 2) more directionally uniform entry distribution than anisotropic one in the vector space. Relevant steps are explained in Section \ref{sec:unlimited} and \ref{sec:pst}. The dictionary resources used for our work are described in Section \ref{sec:datasets}. 

\subsection{Unlimited entries}
\label{sec:unlimited}
Given a dictionary dataset with $n$ entries, to fully exploit the dataset by assigning vector representations to all $n$ entries, we take advantage of the special data structure of dictionary entries. Unlike words or phrases in the context of texts, the semantics of dictionary entries can be represented by their definition sentences. This inspires us to build entry embeddings by means of sentence embeddings. By considering polysemous or paraphrased definitions, for any entry $e$ with $m$ ($\geq 1$) definitions in the dataset, the average pooling of last hidden states of ``[CLS] definition sentence [SEP]'' output from any encoder are taken to represent the entry (formulated in Equation \ref{equ:e1}). For example, if pooling is CLS, after input ``[CLS] give new life or vigor to [SEP]'' and ``[CLS] imbue something with new life and vitality [SEP]'' to BERT, the average of last hidden states of [CLS] is the entry embedding for ``revitalize''. Since various pooling strategies are not the focus of this paper, we use only two basic pooling strategies --- CLS or Mean \citep{sbert} --- for our approach. Corresponding entry embeddings are denoted as \textbf{\emph{AC}} (Average CLS) and \textbf{\emph{AMP}} (Average Mean Pooling) respectively.
\begin{equation}
    \begin{aligned}
     entry\_embed(e) = \frac{1} {m}\sum\limits_{j=0}^{m-1}pooling(encoder([CLS]definition_{j}[SEP]))
    \end{aligned}
\label{equ:e1}
\end{equation}

\subsection{Progressive separate training}
\label{sec:pst}
\begin{figure*}[tp]
\centering
\includegraphics[width=0.8\linewidth]{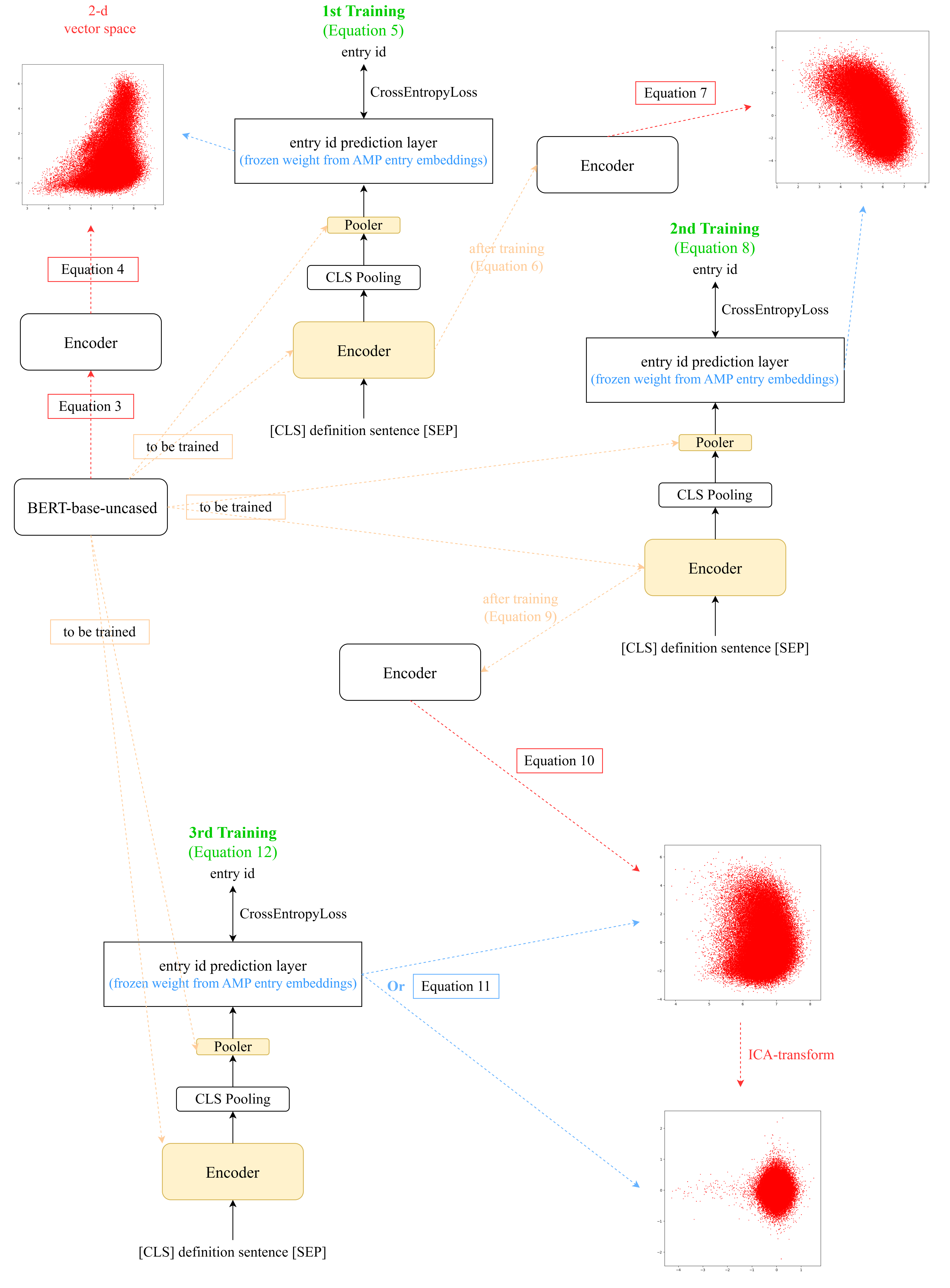}
\vspace{5em}
\caption{Progressive Separate Training (PST) shown by the example of BERT-base-uncased. To echo the gist of our paper, we use the visualization of 2-d vector space to illustrate entry embeddings, where entries are distributed on the two semantic axes with biggest variances. In this example, the pooling used to encode sentence embeddings during training is CLS and entry embeddings are $AMP$, and the total step of PST is three, but they can be different for other models. The used dictionary dataset is $a$ in Table \ref{tab:datasets}.}
\label{fig:pst}
\end{figure*}

Equation \ref{equ:e1} enables us to fully exploit dictionary resources. However, it should be noted that the encoder to build entry embeddings is used before training. For this reason, our embeddings can also suffer from anisotropic vector space if the encoder is a raw pre-trained model. As shown in the upper-left corner of Figure \ref{fig:pst} for BERT-base-uncased model, the SVD visualization of 2-d vector space resembles a narrow cone, which is similar to the one of word embeddings reported in \citet{anisotropicwordembeddingsregularization}'s paper.

Although the encoder trained with the anisotropic vector space (1st Training in Figure \ref{fig:pst}) can already encode sentence embeddings that are better than $DefSent$ (see Section \ref{sec:raw-models}), a pursuit worthy of exploration remains if the vector space is filled with entry distribution more directionally uniform. The most intuitive way is to follow \citet{ICA-transform} to use ICA algorithm to transform the anisotropic vector space, but this is not very effective for our training objective (explained later in this subsection). To further explore the potential of our approach, we devise a novel method named Progressive Separate Training (PST), which can turn the anisotropic vector space progressively into a quasi-isotropic vector space.

Given a dictionary dataset containing entries $e^{'}$, let ($e$, $s$) be any entry-sentence pair in the dataset, and for each individual entry $e \in e^{'}$, it has $m$ ($\geq 1$) definitions\footnote{For any $e$, ($e$, $s$) encompasses the indication of all ($e$, $definition_{j}$), where $j$ is the index within $[0, m-1]$.}. Given an auto-encoding model, let $\theta$ be the parameters of its pooler and encoder, $\theta_{e}$ exclusively be the parameters of its encoder, and $h^{cls}_{\theta}(s)$ denote the representation output from the pooler, where ``cls'' indicates that sentence embeddings are encoded by CLS pooling during training. Equations \ref{equ:1st-frozen-encoder} - \ref{equ:3rd-training}, explain the PST in Figure \ref{fig:pst} for the BERT-base-uncased model.
\begin{equation}
    \begin{aligned}
     encoder \xleftarrow{\text{weight}} \theta_{e}
    \end{aligned}
\label{equ:1st-frozen-encoder}
\end{equation}
\begin{equation}
    \begin{aligned}
     \forall e \in e^{'}, AMP\_entry\_embed(e) = \frac{1} {m}\sum\limits_{j=0}^{m-1}mean\_pooling(encoder([CLS]definition_{j}[SEP]))
    \end{aligned}
\label{equ:1st-e}
\end{equation}
\begin{equation}
    \begin{aligned}
      \theta^{*} = \nabla_{\theta} -log \frac{exp(AMP\_entry\_embed(e) \cdot h^{cls}_{\theta}(s))} {\sum\limits_{e^{'}} exp(AMP\_entry\_embed(e^{'}) \cdot h^{cls}_{\theta}(s))} \text{ , 1-epoch gradient descent}
    \end{aligned}
\label{equ:1st-training}
\end{equation}

\begin{equation}
    \begin{aligned}
     encoder \xleftarrow{\text{weight}} \theta^{*}_{e}
    \end{aligned}
\label{equ:2nd-frozen-encoder}
\end{equation}
\begin{equation}
    \begin{aligned}
     \forall e \in e^{'}, AMP\_entry\_embed(e) = \frac{1} {m}\sum\limits_{j=0}^{m-1}mean\_pooling(encoder([CLS]definition_{j}[SEP]))
    \end{aligned}
\label{equ:2nd-e}
\end{equation}
\begin{equation}
    \begin{aligned}
     \theta^{**} = \nabla_{\theta} -log \frac{exp(AMP\_entry\_embed(e) \cdot h^{cls}_{\theta}(s))} {\sum\limits_{e^{'}} exp(AMP\_entry\_embed(e^{'}) \cdot h^{cls}_{\theta}(s))} \text{ , 1-epoch gradient descent}
    \end{aligned}
\label{equ:2nd-training}
\end{equation}
\begin{equation}
    \begin{aligned}
     encoder \xleftarrow{\text{weight}} \theta^{**}_{e}
    \end{aligned}
\label{equ:3rd-frozen-encoder}
\end{equation}
\begin{equation}
    \begin{aligned}
     \forall e \in e^{'}, AMP\_entry\_embed(e) = \frac{1} {m}\sum\limits_{j=0}^{m-1}mean\_pooling(encoder([CLS]definition_{j}[SEP]))
    \end{aligned}
\label{equ:3rd-e}
\end{equation}
\begin{equation}
    candidate\_entry\_embed = \left\{
\begin{array}{lll}
      AMP\_entry\_embed \\
      ICA-transform(AMP\_entry\_embed) \\
\end{array} 
\right. 
\label{equ:candidate-e}
\end{equation}
\begin{equation}
    \begin{aligned}
     \theta^{***} = \nabla_{\theta} -log \frac{exp(candidate\_entry\_embed(e) \cdot h^{cls}_{\theta}(s))} {\sum\limits_{e^{'}} exp(candidate\_entry\_embed(e^{'}) \cdot h^{cls}_{\theta}(s))} \text{ , 1-epoch gradient descent}
    \end{aligned}
\label{equ:3rd-training}
\end{equation}
\begin{equation}
    \begin{aligned}
     encoder \xleftarrow{\text{weight}} \theta^{***}_{e}
    \end{aligned}
\label{equ:4th-frozen-encoder}
\end{equation}
\begin{equation}
    \begin{aligned}
     sentence\_embed(s) = pooling(encoder([CLS]s[SEP])), pooling = \left\{
\begin{array}{lll}
      CLS \\
      Mean \\
      Prompt \\
\end{array}
\right. 
    \end{aligned}
\label{equ:sentence-e}
\end{equation}

\begin{figure*}[t]
\centering
\includegraphics[width=0.8\linewidth]{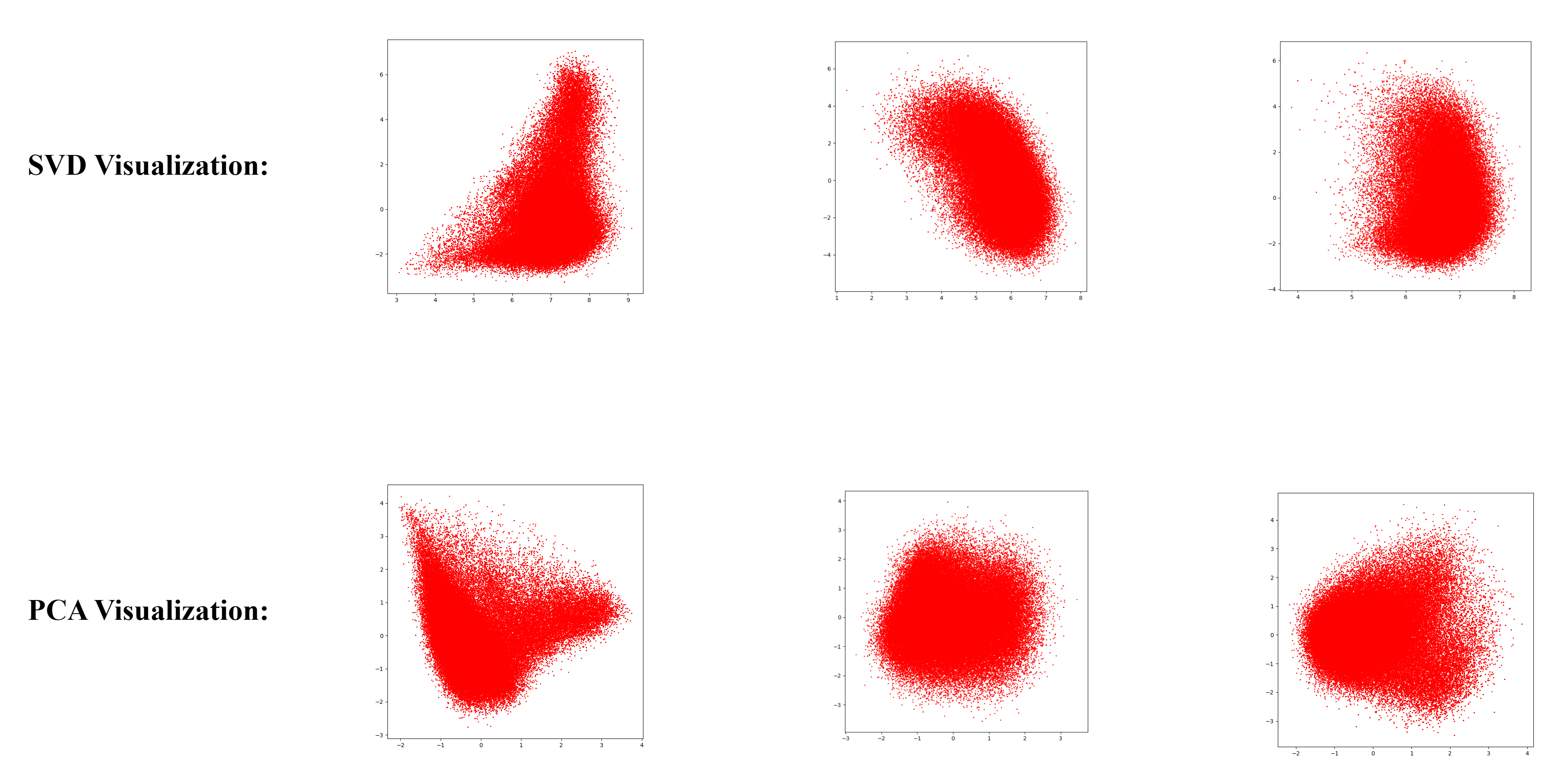}
\vspace{1em}
\caption{The PCA visualizations (with whitening) that correspond to the SVD Visualizations.}
\label{fig:svd-pca}

\vspace*{\floatsep}

\includegraphics[width=0.9\linewidth]{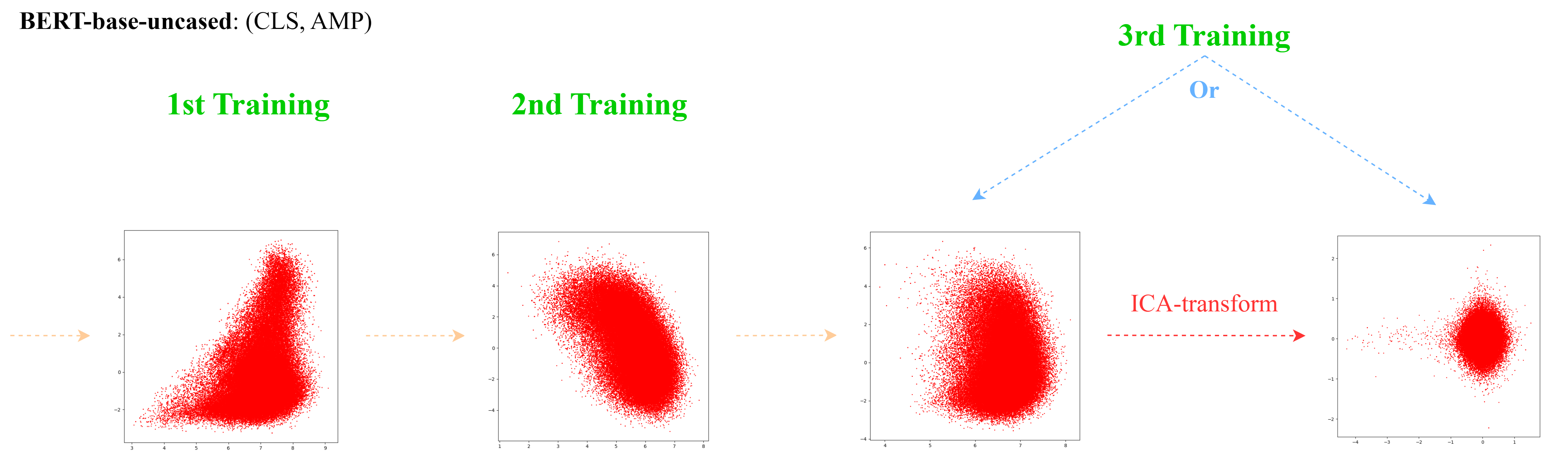}
\vspace{2em}
\caption{Simplified streamline for the PST illustrated by the example of BERT-base-uncased in Figure \ref{fig:pst}. The combination (CLS, AMP) means that during training, the pooling used to encode sentence embeddings is CLS and entry embeddings are $AMP$.}
\label{fig:streamline-bb}
\end{figure*}

As shown in the upper-right corner of Figure \ref{fig:pst}, the encoder after 1st Training can be used to build entry embeddings approaching a rotated whitened space. This is an improvement compared to the previous anisotropic one, but a linear regression line can be drawn to indicate negative correlation. Progressively, the encoder trained with this space (2nd Training in Figure \ref{fig:pst}) can yield entry embeddings resembling a whitened space (the visualization after Equation \ref{equ:3rd-e} in Figure \ref{fig:pst}). This is the space that we address as \textbf{quasi-isotropic}: the entry distribution is more directionally uniform than the anisotrpic one but is not stringently isotropic. The centered entry embeddings\footnote{ PCA visualizations with or without whitening have same shape with different scale. To center the weight matrix of entry embeddings first for SVD visualization is equal to PCA visualization without whitening (divided by square root of eigen-values for unit variances).}, as reflected by the PCA visualizations in Figure \ref{fig:svd-pca}, further shows directionally uniform in a progressive manner (why the 3rd is better than the 2nd is explained in Section \ref{sec:raw-models}). \emph{Since encoders trained with centered entry embeddings or non-centered one can encode sentence embeddings of same quality, plus both SVD and PCA visualizations can illustrate directionally uniform in a progressive manner, we use non-centered entry embeddings and SVD visualizations throughout the paper}.

We further provide another candidate vector space for 3rd Training: \textbf{ICA-transformed}. The ICA-transformed one is the quasi-isotropic space processed by the ICA algorithm\footnote{We do not delineate ICA algorithm in this paper, as it is a common classical procedure of data analysis, aiming to whiten a given vector space first, and then to maximize non-Gaussianity of distributions on semantic axes by cost functions approaching negentropy. For a clear reference, see \citet{ICA}, \citet{ICA-latest-intro} or \citet{ICA-transform}.}, which is more effective for our training objective when compared to the anisotropic space directly processed by ICA. As shown in Figure \ref{fig:svd-pca}, the PCA visualization of the anisotropic distribution is far from sphere-like. Since the prerequisite step of ICA algorithm is to whiten a given space first \citep{ICA}, it is not surprising that the anisotropic distribution processed by the ICA algorithm is less effective than the quasi-isotropic one. The ICA-transformed distribution, with a centered sphere, has typically big skewness and kurtosis (i.e., asymmetric and heavy tails) to emphasize outstanding entries as outliers on different semantic axes. We will show in Section \ref{sec:sts} that the two candidate entry embeddings are effective for training different encoders.

Figure \ref{fig:streamline-bb} illustrates a simplified streamline of the PST in Figure \ref{fig:pst} for BERT-base-uncased model, in order to provide a concise reading. Later in Section \ref{sec:sts}, we will also demonstrate simplified streamlines for other models. 

Two settings, encoding combinations and total step of PST, are flexible. There are four combinations --- (CLS, AMP), (Mean, AMP), (CLS, AC), and (Mean, AC) --- for training different encoders. In the example of BERT-base-uncased in Figures \ref{fig:pst} and \ref{fig:streamline-bb}, the most effective combination is (CLS, AMP): the pooling used to encode sentence embeddings during training is CLS and entry embeddings are $AMP$. The combination is determined in 1st Training of PST through grid search. Regarding the total step of PST, in this example for BERT-base-uncased, the final step is 3rd Training. This is same for other raw pre-trained BERT or RoBERTa models. When $DefSent+$ is applied to further training data-augmented models, the total step of PST is one or two. Excessive separate training will diminish the quality of sentence embeddings.

Another two settings are noteworthy. For each training step during PST, the encoder trained by the previous step should not be used for training in the next step; e.g., to replace the $\theta$ of $h^{cls}_{\theta}(s)$ with $\theta^{*}$ in Equation \ref{equ:2nd-training} for 2nd Training of BERT-base-uncased model. Otherwise, it would be tantamount to training an encoder for more than 1 epoch, which will lessen the quality of sentence embeddings. The reason why this approach requires only 1-epoch training is not explained in \citet{defsent}'s paper, but we will share our insight in Appendix \ref{sec:hyperparameter} describing hyperparameters. This is the meaning of ``Separate'' in the name of PST --- parameters of encoders from original models are separately optimized at different steps of PST. Regarding the pooler, the weight of dense layer in RobertaPooler carries insufficient semantic information due to no NSP pre-training objective. We resolve it by copying the weight data of dense layer in MLM head to the dense layer in RobertaPooler, and also change the activation function from Tanh to GELU correspondingly (BertPooler remains as it is). After training, as our goal is to encode sentence embeddings, the optimized parameters of pooler are no longer useful, but still valuable if one wants to implement a reverse dictionary.   

The encoder after the final step of PST --- e.g., Equation \ref{equ:4th-frozen-encoder} for BERT-base-uncased model after 3rd Training of PST --- is used to encode sentence embeddings for any input sentence $s$ in Equation \ref{equ:sentence-e}. We also attempt the Prompt pooling \citep{promptbert,sncse}: the last hidden states of [MASK] token for input like ``This sentence: `New York is a big city.' means [MASK].'' ($\langle$mask$\rangle$ for RoBERTa) to represent sentence embeddings.

The strength of PST is that it can develop the vector space of entry embeddings naturally into an optimal one catering for different encoders to be trained. We will conclude this finding in Section \ref{sec:conclusion} based on the experimental results presented in Sections \ref{sec:sts} and \ref{sec:senteval}.

\subsection{Dictionary datasets \& usage}
\label{sec:datasets}
The datasets are summarized in Table \ref{tab:datasets}, which are formed from Oxford released by \citet{dictionary} and WordNet \citep{wordnet} in an incremental manner. The basic one, Dataset $c$, follows the setting of $DefSent$ \citep{defsent}, consisting of single words only in the Oxford resource based on BERT vocabulary\footnote{The number of single words in the Oxford resource based on BERT and RoBERTa vocabulary nearly the same reported in $DefSent$ \citep{defsent}, but target entry ids for all entries are same for BERT and RoBERTa in our work. Considering this, only single words from BERT vocabulary are used to emphasize basic entry numbers.}. Two settings, not implementable for $DefSent$, are also investigated: Dataset $b$, a fully-exploited Oxford resource; Dataset $a$, a combination with other resources like WordNet.

\begin{table*}[t]
\centering
\caption{Overview of dictionary datasets. In Dataset $a$, duplicate entries from different resources are combined, and duplicate definitions for each entry are removed if exist.}
\label{tab:datasets}
\resizebox{0.7\linewidth}{!}{%
\begin{tabular}{ccccc}
\hline
Dataset index &
Source &
Definitions &
Entries &
\begin{tabular}[c]{@{}c@{}}Max (Median) length\\ (BERT / RoBERTa)\end{tabular} \\ \hline
a & Oxford + WordNet & 331,472 & 150,518 & 138 (12) / 141 (12)  \\ 
b & Oxford & 122,209 & 36,767   & 108 (12) / 113 (12) \\ 
c & \begin{tabular}[c]{@{}c@{}}Oxford\\ (single words only)\end{tabular} & 67,553 & 9,647   & 85 (12) / 93 (12) \\ \hline
\end{tabular}%
}
\end{table*}

As explained in Section \ref{sec:motivation}, $n$ and $\mathbb{d}$ determine the quality of entry embeddings for definition sentences to project into. Later in Section \ref{sec:raw-models}, we will perform 1st Training of PST for all encoders using Datasets $a$, $b$, and $c$, to show the difference made by $n$. As Dataset $a$ generates best results, we will perform full PST for all encoders using the $a$ to show the difference made by $\mathbb{d}$.

\section{Unsupervised semantic textual similarity tasks}
\label{sec:sts}
STS tasks without using any training data are used to test the capability of our sentence embeddings for the major application\footnote{Following DefSent \citep{defsent}, we use the test datasets under /\underline{\textbf{data/downstream}} of SentEval \citep{senteval}.}. The tasks cover STS 2012-2016 \citep{sts12,sts13,sts14,sts15,sts16}, STS benchmark \citep{sts-b}, and SICK-Relatedness \citep{sick-r}. The metric is Spearman's rank correlation $\rho$ × 100 between cosine similarities of sentence embeddings and human ratings.

We follow \citet{defsent} to perform training for each encoder with 10 runs of different random seeds. There are two types of result: statistical result (mean and standard deviation of 10 runs) and best result (highest average score achieved by the optimal enoder out of 10 runs). For a fair comparison, statistical results are reported in Section \ref{sec:raw-models} for the comparison with $DefSent$, and we leave corresponding best results in Appendix \ref{sec:supplementary-result}. For the comparison with data-augmented models in Section \ref{sec:data-augmented-models}, as the results reported in their papers are achieved by the released models, we report best results for comparison, and leave corresponding statistical results in Appendix \ref{sec:supplementary-result}. For both types of result, they are based on the best pooling strategy attempted from Mean, CLS, or Prompt, which are also summarized in Appendix \ref{sec:supplementary-result}.

Since there are 10 runs of training, for the progressive steps of PST, the optimal encoder from the previous step is used to build entry embeddings for the next step. For example, the $\theta^{*}_{e}$ in Equation \ref{equ:2nd-frozen-encoder} is from the optimal model out of 10 runs after 1st Training of PST for BERT-base-uncased.

The nomenclature of $DefSent+$ encoders is $DefSent+^{PST-(pooling, type)-index}_{model\_name}$; e.g., $DefSent+^{1st-(CLS, AMP)-a}_{BERT-base}$ means that the experimental results are achieved by a BERT-base-uncased model trained from 1st Training of PST using Dataset $a$, and the encoding combination used during training is (CLS, $AMP$). After the final step of PST, for instance, the ``ICA'' in the lower brackets of $DefSent+^{3rd-(CLS, AMP)-a}_{BERT-base(ICA)}$ indicate that results are from training with the ICA-transformed vector space. If a ``$*$'' is attached to an encoder, the encoder will be released by us and also will be tested in Section \ref{sec:senteval} for the capability of feature-based transfer.

\begin{table*}[t]
\centering
\caption{The statistical results out of 10 runs are reported for $DefSent+$ encoders trained from raw pre-trained models. The baselines are excerpted from $DefSent$ \citep{defsent}, which are based on their best pooling results.}
\label{tab:raw-models}
\resizebox{1.0\linewidth}{!}{%
\begin{tabular}{lcccccccc}
\hline
Encoder & STS12 & STS13 & STS14 & STS15 & STS16 & STS-B & SICK-R & Avg. \\ \hline
\begin{tabular}[l]{@{}l@{}}$DefSent_{BERT-base}$\end{tabular}     & 67.30\scriptsize$\pm$0.26 & 81.96\scriptsize$\pm$0.24 & 71.92\scriptsize$\pm$0.28 & 77.68\scriptsize$\pm$0.47 & 76.71\scriptsize$\pm$0.48 & 76.90\scriptsize$\pm$0.40 & 73.28\scriptsize$\pm$0.30 & 75.11\scriptsize$\pm$0.21  \\ 
                                                                                    \begin{tabular}[l]{@{}l@{}}$DefSent_{BERT-large}$\end{tabular}
                                                                                    & 64.18\scriptsize$\pm$0.96 & 82.76\scriptsize$\pm$0.42 & 73.14\scriptsize$\pm$0.32 & 79.66\scriptsize$\pm$0.92 & 77.93\scriptsize$\pm$0.78 & 77.89\scriptsize$\pm$0.89 & 73.98\scriptsize$\pm$0.46 & 75.65\scriptsize$\pm$0.53 \\
                                                                                    \begin{tabular}[l]{@{}l@{}}$DefSent_{RoBERTa-base}$\end{tabular}
                                                                                    & 65.55\scriptsize$\pm$0.89 & 80.84\scriptsize$\pm$0.26 & 71.87\scriptsize$\pm$0.39 & 78.77\scriptsize$\pm$0.70 & 79.29\scriptsize$\pm$0.27 & 78.13\scriptsize$\pm$0.61 & 74.92\scriptsize$\pm$0.18 & 75.62\scriptsize$\pm$0.38 \\
                                                                                    \begin{tabular}[l]{@{}l@{}}$DefSent_{RoBERTa-large}$\end{tabular}
                                                                                    & 62.89\scriptsize$\pm$1.42 & 77.99\scriptsize$\pm$1.88 & 69.83\scriptsize$\pm$1.66 & 75.60\scriptsize$\pm$1.51 & 79.63\scriptsize$\pm$0.60 & 79.34\scriptsize$\pm$0.48 & 74.04\scriptsize$\pm$0.84 & 74.19\scriptsize$\pm$0.88 \\ \hline
\begin{tabular}[l]{@{}l@{}}$DefSent+^{1st-(Mean, AMP)-c}_{BERT-base}$\end{tabular}    & 65.24\scriptsize$\pm$0.53 & 82.25\scriptsize$\pm$0.24 & 72.37\scriptsize$\pm$0.25 & 80.80\scriptsize$\pm$0.48 & 77.11\scriptsize$\pm$0.36 & 77.38\scriptsize$\pm$0.43 & 72.38\scriptsize$\pm$0.29 & 75.36\scriptsize$\pm$0.22  \\ 
                                                                                    \begin{tabular}[l]{@{}l@{}}$DefSent+^{1st-(Mean, AMP)-c}_{BERT-large}$\end{tabular}
                                                                                    & 66.28\scriptsize$\pm$0.63 & 84.17\scriptsize$\pm$0.34 & 75.25\scriptsize$\pm$0.42 & 82.72\scriptsize$\pm$0.35 & 78.87\scriptsize$\pm$0.45 & 78.85\scriptsize$\pm$0.52 & 74.37\scriptsize$\pm$0.38 & 77.22\scriptsize$\pm$0.27 \\
                                                                                    \begin{tabular}[l]{@{}l@{}}$DefSent+^{1st-(CLS, AC)-c}_{RoBERTa-base}$\end{tabular}
                                                                                    & 66.70\scriptsize$\pm$0.40 & 81.23\scriptsize$\pm$0.47 & 72.36\scriptsize$\pm$0.34 & 81.15\scriptsize$\pm$0.26 & 79.67\scriptsize$\pm$0.31 & 77.87\scriptsize$\pm$0.53 & 72.49\scriptsize$\pm$0.58 & 75.92\scriptsize$\pm$0.19 \\
                                                                                    \begin{tabular}[l]{@{}l@{}}$DefSent+^{1st-(CLS, AC)-c}_{RoBERTa-large}$\end{tabular}
                                                                                    & 67.32\scriptsize$\pm$1.07 & 83.61\scriptsize$\pm$0.61 & 74.96\scriptsize$\pm$0.52 & 83.12\scriptsize$\pm$0.51 & 79.09\scriptsize$\pm$0.59 & 80.23\scriptsize$\pm$0.55 & 73.40\scriptsize$\pm$0.67 & 77.39\scriptsize$\pm$0.56 \\ \hdashline
\begin{tabular}[l]{@{}l@{}}$DefSent+^{1st-(CLS, AMP)-b}_{BERT-base}$\end{tabular}    & 68.03\scriptsize$\pm$0.64 & 81.97\scriptsize$\pm$0.68 & 74.37\scriptsize$\pm$0.66 & 81.59\scriptsize$\pm$0.41 & 79.18\scriptsize$\pm$0.48 & 79.48\scriptsize$\pm$0.55 & 71.64\scriptsize$\pm$0.61 & 76.61\scriptsize$\pm$0.38  \\ 
                                                                                    \begin{tabular}[l]{@{}l@{}}$DefSent+^{1st-(CLS, AMP)-b}_{BERT-large}$\end{tabular}
                                                                                    & 69.79\scriptsize$\pm$1.21 & 84.56\scriptsize$\pm$0.89 & 77.32\scriptsize$\pm$0.49 & 83.81\scriptsize$\pm$0.36 & 81.13\scriptsize$\pm$0.24 & 81.15\scriptsize$\pm$0.40 & 73.41\scriptsize$\pm$0.52 & 78.74\scriptsize$\pm$0.39 \\
                                                                                    \begin{tabular}[l]{@{}l@{}}$DefSent+^{1st-(Mean, AC)-b}_{RoBERTa-base}$\end{tabular}
                                                                                    & 71.25\scriptsize$\pm$0.30 & 81.33\scriptsize$\pm$0.24 & 72.86\scriptsize$\pm$0.27 & 81.74\scriptsize$\pm$0.23 & 80.58\scriptsize$\pm$0.36 & 79.77\scriptsize$\pm$0.15 & 72.91\scriptsize$\pm$0.40 & 77.21\scriptsize$\pm$0.11 \\
                                                                                    \begin{tabular}[l]{@{}l@{}}$DefSent+^{1st-(Mean, AC)-b}_{RoBERTa-large}$\end{tabular}
                                                                                    & 69.65\scriptsize$\pm$0.24 & 84.87\scriptsize$\pm$0.36 & 76.74\scriptsize$\pm$0.32 & 83.99\scriptsize$\pm$0.30 & 81.49\scriptsize$\pm$0.19 & 81.80\scriptsize$\pm$0.27 & 74.43\scriptsize$\pm$0.40 & 79.00\scriptsize$\pm$0.18 \\ \hdashline
\begin{tabular}[l]{@{}l@{}}$DefSent+^{1st-(CLS, AMP)-a}_{BERT-base}$\end{tabular}    & 69.09\scriptsize$\pm$0.61 & 83.44\scriptsize$\pm$0.32 & 75.88\scriptsize$\pm$0.36 & 82.63\scriptsize$\pm$0.40 & 80.26\scriptsize$\pm$0.34 & 81.13\scriptsize$\pm$0.34 & 72.73\scriptsize$\pm$0.42 &  77.88\scriptsize$\pm$0.22 \\ 
                                                                                    \begin{tabular}[l]{@{}l@{}}$DefSent+^{1st-(CLS, AMP)-a}_{BERT-large}$\end{tabular}
                                                                                    & 71.29\scriptsize$\pm$1.08 & 84.81\scriptsize$\pm$1.29 & 77.66\scriptsize$\pm$1.06 & 83.84\scriptsize$\pm$0.91 & 81.14\scriptsize$\pm$0.90 & 82.55\scriptsize$\pm$0.79 & 74.50\scriptsize$\pm$0.80 & 79.40\scriptsize$\pm$0.72 \\
                                                                                    \begin{tabular}[l]{@{}l@{}}$DefSent+^{1st-(Mean, AC)-a}_{RoBERTa-base}$\end{tabular}
                                                                                    & 72.23\scriptsize$\pm$0.28 & 82.79\scriptsize$\pm$0.20 & 74.96\scriptsize$\pm$0.22 & 82.87\scriptsize$\pm$0.13 & 81.18\scriptsize$\pm$0.21 & 81.22\scriptsize$\pm$0.23 & 73.87\scriptsize$\pm$0.19 & 78.44\scriptsize$\pm$0.14 \\
                                                                                    \begin{tabular}[l]{@{}l@{}}$DefSent+^{1st-(Mean, AC)-a}_{RoBERTa-large}$\end{tabular}
                                                                                    & 70.83\scriptsize$\pm$0.61 & 85.87\scriptsize$\pm$0.21 & 77.61\scriptsize$\pm$0.29 & 84.33\scriptsize$\pm$0.32 & 82.06\scriptsize$\pm$0.41 & 82.43\scriptsize$\pm$0.37 & 75.48\scriptsize$\pm$0.20 & 79.81\scriptsize$\pm$0.19 \\ \hline
\begin{tabular}[l]{@{}l@{}}$DefSent+^{2nd-(CLS, AMP)-a}_{BERT-base}$\end{tabular}    & 70.18\scriptsize$\pm$0.38 & 84.18\scriptsize$\pm$0.30 & 76.54\scriptsize$\pm$0.36 & 83.21\scriptsize$\pm$0.40 & 80.66\scriptsize$\pm$0.27 & 81.50\scriptsize$\pm$0.40 & 72.53\scriptsize$\pm$0.71 &  78.40\scriptsize$\pm$0.23 \\ 
                                                                                    \begin{tabular}[l]{@{}l@{}}$DefSent+^{2nd-(CLS, AMP)-a}_{BERT-large}$\end{tabular}
                                                                                    & 69.52\scriptsize$\pm$0.77 & 85.87\scriptsize$\pm$0.64 & 77.46\scriptsize$\pm$0.63 & 82.95\scriptsize$\pm$0.83 & 80.54\scriptsize$\pm$0.79 & 80.99\scriptsize$\pm$0.59 & 74.45\scriptsize$\pm$0.77 & 78.83\scriptsize$\pm$0.33 \\
                                                                                    \begin{tabular}[l]{@{}l@{}}$DefSent+^{2nd-(Mean, AC)-a}_{RoBERTa-base}$\end{tabular}
                                                                                    & 72.55\scriptsize$\pm$0.34 & 84.33\scriptsize$\pm$0.32 & 76.12\scriptsize$\pm$0.41 & 83.39\scriptsize$\pm$0.30 & 81.11\scriptsize$\pm$0.37 & 80.96\scriptsize$\pm$0.43 & 73.46\scriptsize$\pm$0.30 & 78.85\scriptsize$\pm$0.25 \\
                                                                                    \begin{tabular}[l]{@{}l@{}}$DefSent+^{2nd-(Mean, AC)-a}_{RoBERTa-large}$\end{tabular}
                                                                                    & 72.98\scriptsize$\pm$0.50 & 84.20\scriptsize$\pm$0.70 & 77.54\scriptsize$\pm$0.67 & 84.29\scriptsize$\pm$0.40 & 82.09\scriptsize$\pm$0.50 & 83.08\scriptsize$\pm$0.44 & 75.29\scriptsize$\pm$0.52 & 79.96\scriptsize$\pm$0.31 \\ \hline
\begin{tabular}[l]{@{}l@{}}$DefSent+^{3rd-(CLS, AMP)-a}_{BERT-base}$\end{tabular}    & 69.91\scriptsize$\pm$0.37 & 84.19\scriptsize$\pm$0.23 & 76.90\scriptsize$\pm$0.24 & 83.52\scriptsize$\pm$0.18 & 80.60\scriptsize$\pm$0.17 & 81.17\scriptsize$\pm$0.14 & 72.64\scriptsize$\pm$0.45 &  78.42\scriptsize$\pm$0.16 \\ 
                                                                                    \begin{tabular}[l]{@{}l@{}}$DefSent+^{3rd-(CLS, AMP)-a}_{BERT-large}$\end{tabular}
                                                                                    & 71.48\scriptsize$\pm$0.90 & 85.83\scriptsize$\pm$0.35 & 78.44\scriptsize$\pm$0.55 & 83.52\scriptsize$\pm$0.52 & 81.71\scriptsize$\pm$0.27 & 82.43\scriptsize$\pm$0.32 & 75.04\scriptsize$\pm$0.73 & 79.78\scriptsize$\pm$0.22 \\
                                                                                    \begin{tabular}[l]{@{}l@{}}*$DefSent+^{3rd-(Mean, AC)-a}_{RoBERTa-base}$\end{tabular}
                                                                                    & 72.51\scriptsize$\pm$0.21 & 84.83\scriptsize$\pm$0.14 & 76.70\scriptsize$\pm$0.20 & 83.40\scriptsize$\pm$0.10 & 80.67\scriptsize$\pm$0.28 & 80.95\scriptsize$\pm$0.18 & 74.29\scriptsize$\pm$0.16 & 79.04\scriptsize$\pm$0.12 \\
                                                                                    \begin{tabular}[l]{@{}l@{}}*$DefSent+^{3rd-(Mean, AC)-a}_{RoBERTa-large}$\end{tabular}
                                                                                    & 69.71\scriptsize$\pm$0.53 & 85.54\scriptsize$\pm$0.30 & 78.77\scriptsize$\pm$0.37 & 85.26\scriptsize$\pm$0.28 & 82.71\scriptsize$\pm$0.26 & 82.74\scriptsize$\pm$0.32 & 76.77\scriptsize$\pm$0.29 & 80.22\scriptsize$\pm$0.21 \\ \hdashline
\begin{tabular}[l]{@{}l@{}}*$DefSent+^{3rd-(CLS, AMP)-a}_{BERT-base(ICA)}$\end{tabular}    & 71.23\scriptsize$\pm$0.21 & 84.56\scriptsize$\pm$0.19 & 76.93\scriptsize$\pm$0.22 & 83.19\scriptsize$\pm$0.20 & 80.69\scriptsize$\pm$0.22 & 81.57\scriptsize$\pm$0.28 & 72.38\scriptsize$\pm$0.37 &  78.65\scriptsize$\pm$0.15 \\ 
                                                                                    \begin{tabular}[l]{@{}l@{}}*$DefSent+^{3rd-(CLS, AMP)-a}_{BERT-large(ICA)}$\end{tabular}
                                                                                    & 72.47\scriptsize$\pm$0.46 & 86.37\scriptsize$\pm$0.50 & 78.79\scriptsize$\pm$0.39 & 83.93\scriptsize$\pm$0.39 & 81.76\scriptsize$\pm$0.28 & 82.16\scriptsize$\pm$0.24 & 74.56\scriptsize$\pm$0.43 & 80.00\scriptsize$\pm$0.17 \\
                                                                                    \begin{tabular}[l]{@{}l@{}}$DefSent+^{3rd-(Mean, AC)-a}_{RoBERTa-base(ICA)}$\end{tabular}
                                                                                    & 73.30\scriptsize$\pm$0.38 & 84.59\scriptsize$\pm$0.41 & 75.21\scriptsize$\pm$0.67 & 82.42\scriptsize$\pm$0.46 & 80.19\scriptsize$\pm$0.40 & 80.17\scriptsize$\pm$0.27 & 73.89\scriptsize$\pm$0.39 & 78.54\scriptsize$\pm$0.20 \\
                                                                                    \begin{tabular}[l]{@{}l@{}}$DefSent+^{3rd-(Mean, AC)-a}_{RoBERTa-large(ICA)}$\end{tabular}
                                                                                    & 73.83\scriptsize$\pm$0.25 & 85.38\scriptsize$\pm$0.34 & 77.83\scriptsize$\pm$0.40 & 84.15\scriptsize$\pm$0.41 & 81.68\scriptsize$\pm$0.51 & 81.35\scriptsize$\pm$0.28 & 76.17\scriptsize$\pm$0.18 & 80.05\scriptsize$\pm$0.16 \\ \hline
\end{tabular}%
}
\end{table*}

\subsection{DefSent+ for raw pre-trained models}
\label{sec:raw-models}
In this section, we apply $DefSent+$ to raw pre-trained models, and the experimental results of STS tasks are reported in Table \ref{tab:raw-models}. There are four performance groups separated by solid lines. The first group shows baselines excerpted from $DefSent$ based on their best pooling results. The second group consists of $DefSent+$ encoders optimized at 1st Training of PST using Datasets $c$, $b$, and $a$, so as to show the improvement made by full exploitation of dictionary resources. As encoders trained by Dataset $a$ generate best results, the third and fourth groups present the rest results of full PST, in order to show the difference made by different vector spaces.

\begin{figure*}[t]
\centering
\includegraphics[width=0.8\linewidth]{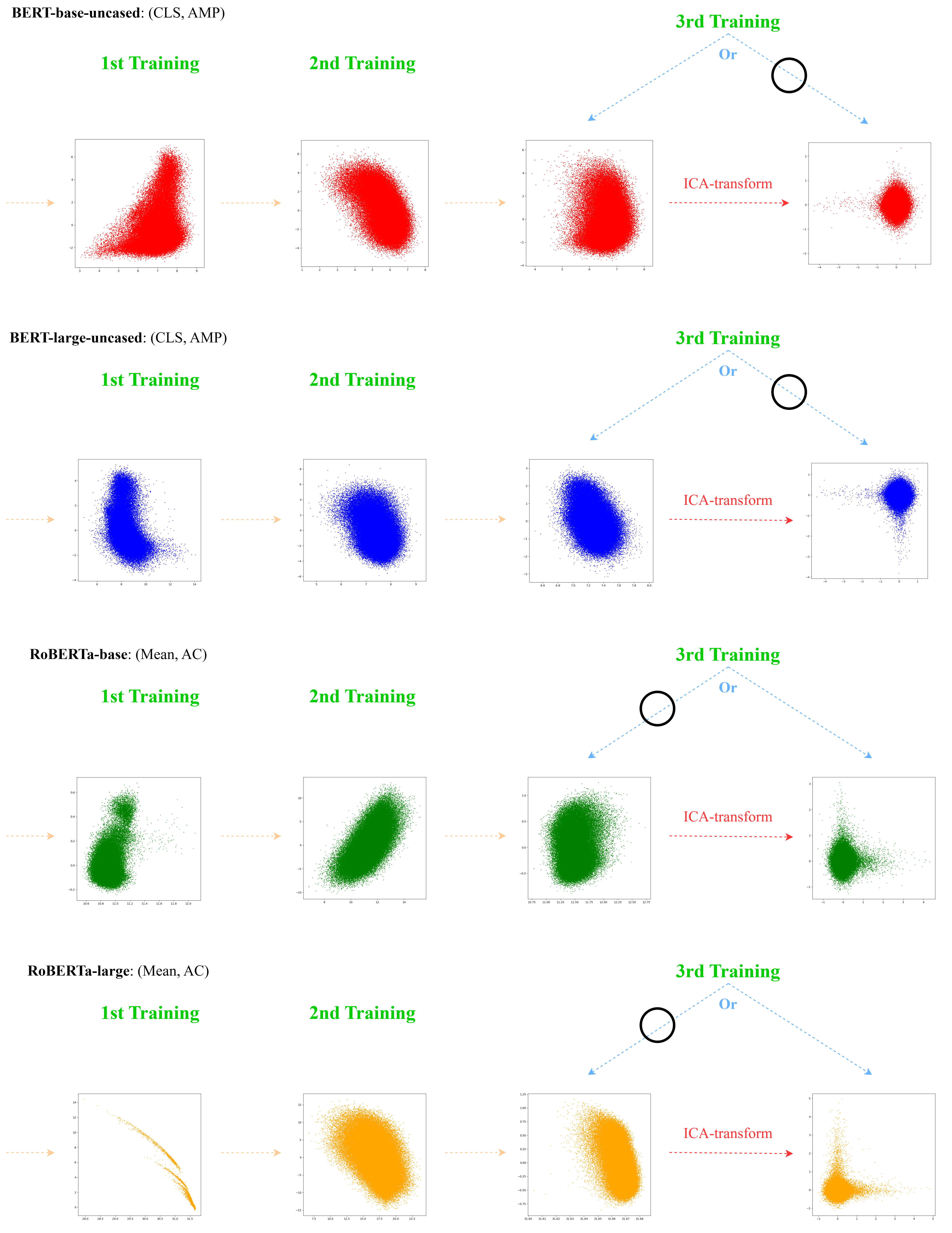}
\vspace{3em}
\caption{Simplified PST streamlines for four raw pre-trained models based on Dataset $a$. The circles indicate the most effective vector spaces for training encoders, and corresponding results are shown in Table \ref{tab:raw-models}.}
\label{fig:pst-raw-models}
\end{figure*}

As shown in the second group of Table \ref{tab:raw-models}, the performance of measuring semantic similarities of sentence pairs is proportional to the size of datasets used to train different encoders. Extrapolating from the increment by comparing $c$ with $b$ and $a$, the essence of fully exploiting any given dictionary resource is validated. With respect to Dataset $c$, the basic resource setting, $DefSent+$ also outperforms $DefSent$ and is comparatively stable, because the only standard deviation over 1 is from $DefSent+^{1st-(CLS, AC)-c}_{RoBERTa-large}$ on STS12. This further validates the effectiveness of Equation \ref{equ:e1} to build entry embeddings, as in this phase of training, our entry embeddings could also suffer from anisotropic vector spaces, but the improvement on performance corroborates that our entry embeddings are relatively effective for this approach. The grid search to determine encoding combinations exemplified by Dataset $a$ is provided in Appendix \ref{sec:supplementary-result}.

\begin{figure*}[t]
\centering
\includegraphics[width=0.8\linewidth]{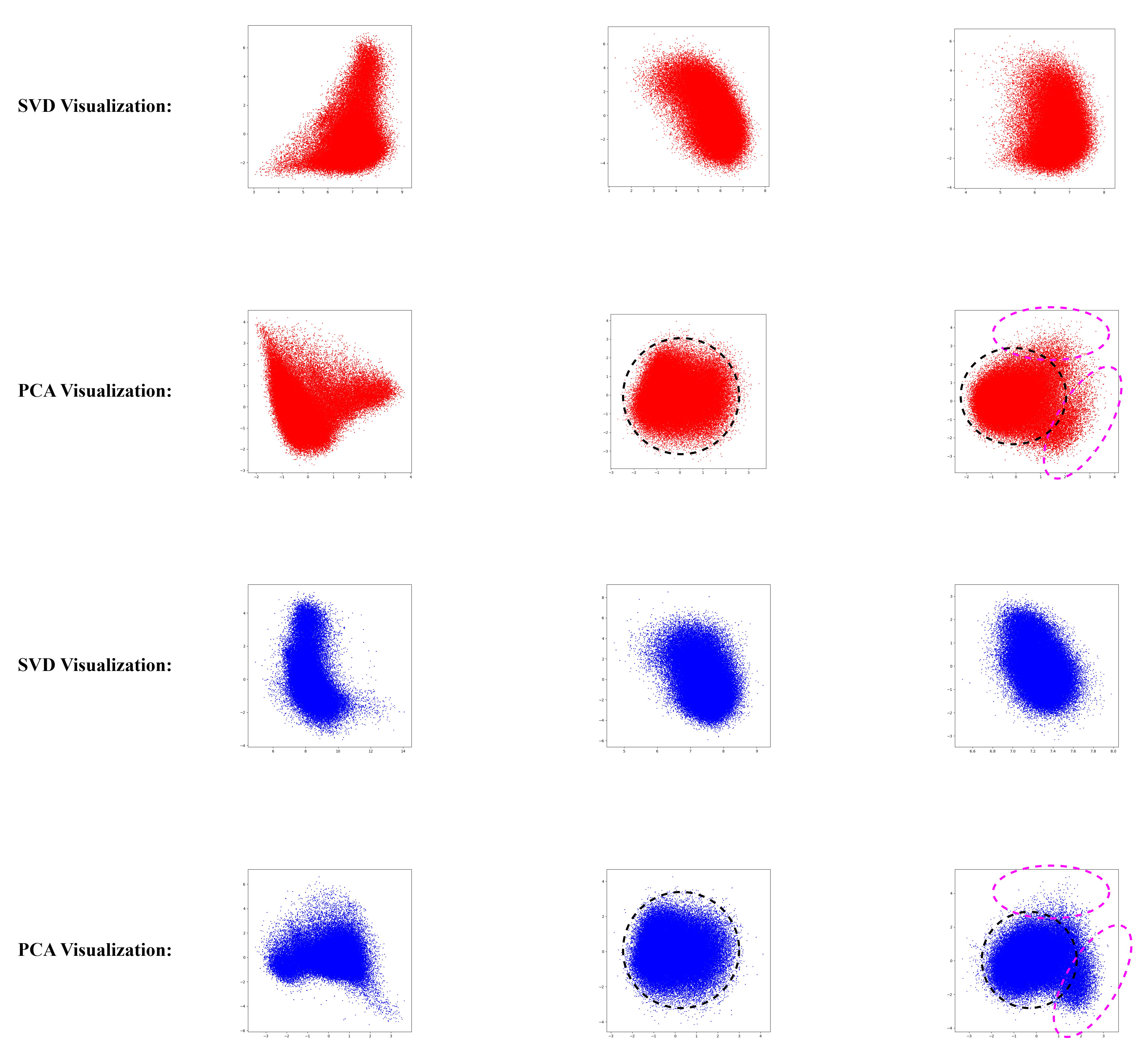}
\vspace{3em}
\caption{The PCA visualizations (with whitening) that correspond to the SVD Visualizations for BERT-base-uncased and BERT-large-uncased. The black dotted circles highlight ideal spheres while the purple dotted circles highlight the entries protruding from the centered sphere.}
\label{fig:svd-pca-BL}
\end{figure*}  

The third and fourth groups present the rest results of full PST based on Dataset $a$, and the corresponding simplified PST streamlines are shown in Figure \ref{fig:pst-raw-models}. As shown in Figure \ref{fig:pst-raw-models}, for the vector spaces used at 1st Training, except for the anisotropic one built from raw BERT-base-uncased, the other three spaces exhibit only the aperture of a cone. Progressively into 2nd Training of PST, the vector spaces used for training resemble rotated ellipses --- entries distributed in the spaces are scattered along all directions when compared to their distributions in the spaces used for 1st Training. When used for 3rd Training, the vector spaces progress into quasi-isotropic distributions, but it seems that for BERT-large-uncased and RoBERTa-large, the ellipse-like distributions are still somewhat rotated without being parallel to semantic axes. However, they are the optimal vector spaces catering for training these two encoders.

\begin{figure*}[t]
\centering
\includegraphics[width=0.8\linewidth]{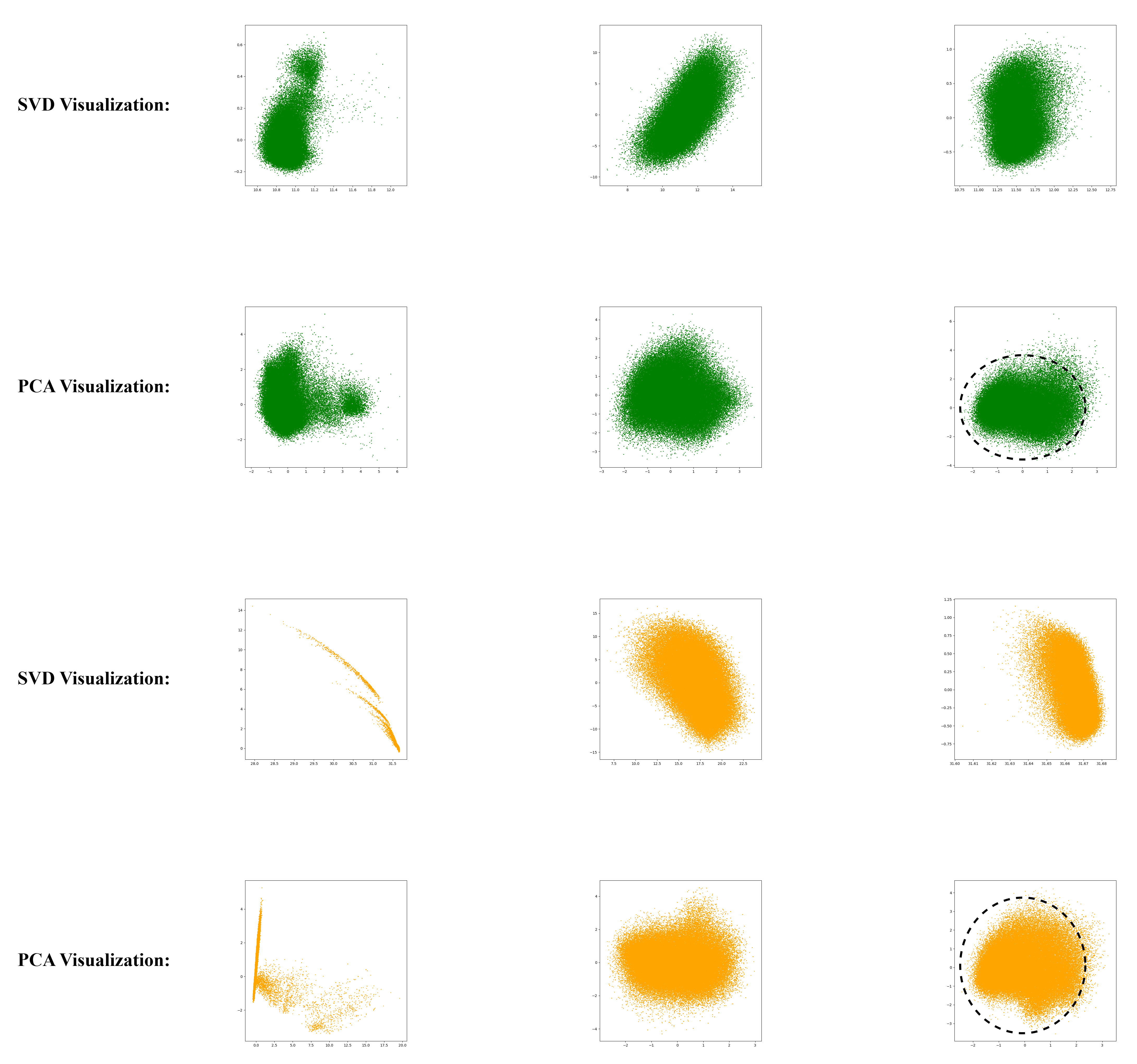}
\vspace{2em}
\caption{The PCA visualizations (with whitening) that correspond to the SVD Visualizations for RoBERTa-base and RoBERTa-large. The black dotted circles highlight the ideal spheres.}
\label{fig:svd-pca-RB-RL}
\end{figure*}

As for BERT-large-uncased, unlike the other raw pre-trained models with performance consistently increasing as vector space evolves, the overall performance on STS tasks of $DefSent+^{2nd-(CLS, AMP)-a}_{BERT-large}$ is lower than the one of $DefSent+^{1st-(CLS, AMP)-a}_{BERT-large}$, albeit $DefSent+^{3rd-(CLS, AMP)-a}_{BERT-large}$, as usual, achieves higher performance than 1st and 2nd. Nevertheless, for encoders relying on isotropic (ICA-transformed) vector space to achieve final highest results, 2nd Training of PST is only a transition step, in our case, for BERT-base-uncased and BERT-large-uncased. The corresponding PCA visualizations for BERT-base-uncased and BERT-large-uncased are illustrated in Figure \ref{fig:svd-pca-BL}. Although $DefSent+^{2nd-(CLS, AMP)-a}_{BERT-large}$ is not better than $DefSent+^{1st-(CLS, AMP)-a}_{BERT-large}$, and the vector space built by it for 3rd Training is still rotated, the vector space after whitening (the lower-right one in Figure \ref{fig:svd-pca-BL}) \textbf{as the prerequisite step for ICA algorithm}, exhibit the similar protruding patterns as the one of BERT-base-uncased in the same figure. We argue that this is the main reason why BERT encoders achieve highest average results on STS tasks with ICA-transformed vector spaces used for 3rd Training. This can be justified by three observations. First, after ICA-transforming the vector spaces used for 2nd Training, the performance on STS tasks for $DefSent+^{2nd-(CLS, AMP)-a}_{BERT-base(ICA)}$ and $DefSent+^{2nd-(CLS, AMP)-a}_{BERT-large(ICA)}$ is 78.45 and 79.42 (not reported in the table for readability), which stay similar level as the best of previous two steps of PST --- 78.42 for $DefSent+^{2nd-(CLS, AMP)-a}_{BERT-base}$ and 79.40 for $DefSent+^{1st-(CLS, AMP)-a}_{BERT-large}$. As reflected by the vertically middle PCA visualizations in Figure \ref{fig:svd-pca-BL}, the entry distributions fall within the ideal spheres for both vector spaces used for 2nd Training, which do not have the same protruding patterns as the vector spaces used for 3rd Training. Second, as shown in Figure \ref{fig:svd-pca-RB-RL} for the PCA visualizations of RoBERTa-base and RoBERTa-large, the vector spaces used for 3rd Training contain entries distributed almost in the ideal spheres, where the protruding patterns are also not observable. As a matter of fact, RoBERTa encoders achieve highest average results on STS tasks by using quasi-isotropic vector spaces for 3rd Training. We leave the third argument in the last paragraph of this subsection, where excessive separate training of PST is described.

\begin{figure*}[t]
\centering
\includegraphics[width=0.9\linewidth]{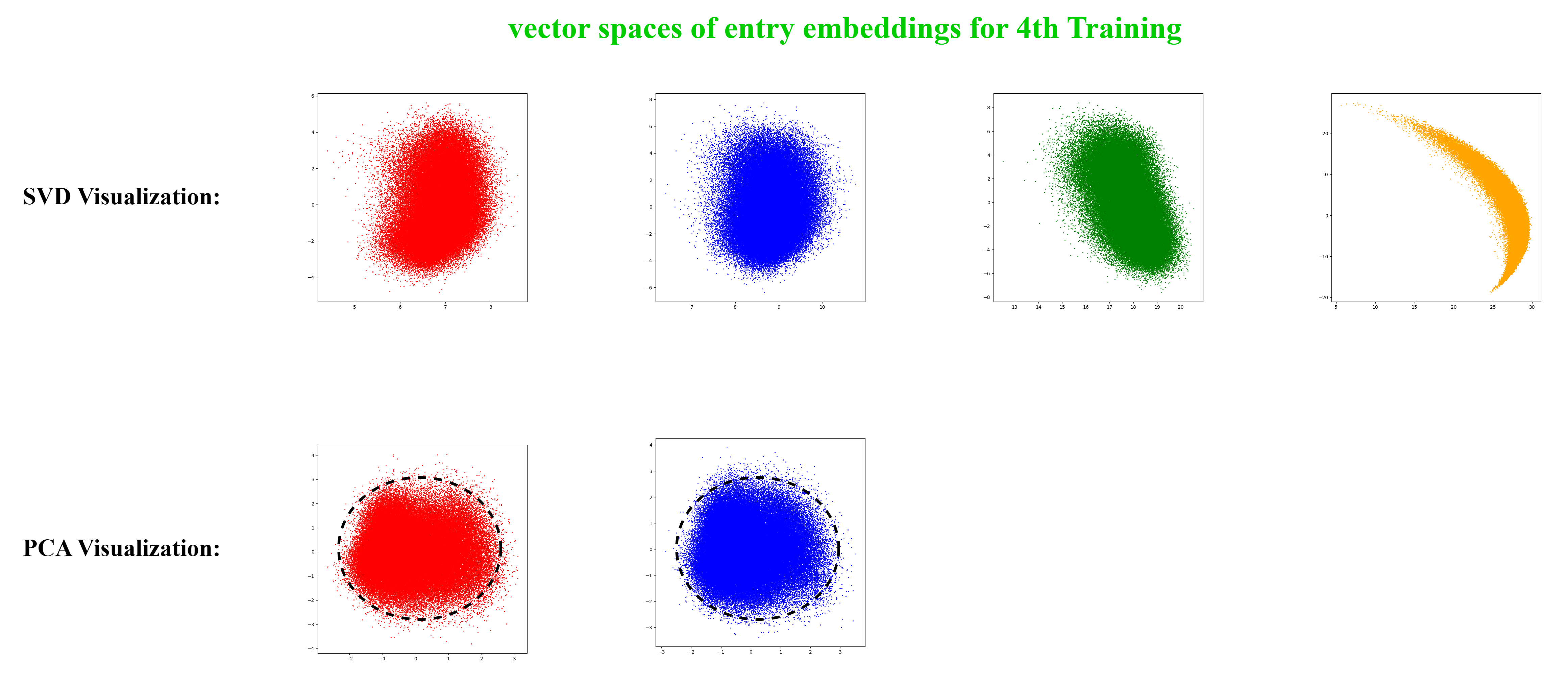}
\vspace{2em}
\caption{The vector spaces of entry embeddings built from encoders after 3rd Training of PST for 4th Training. The black dotted circles highlight the ideal spheres.}
\label{fig:raw-4th}
\end{figure*}

As for RoBERTa-large, same as RoBERTa-base, the encoders after 3rd Training of PST is more powerful than the ones after 1st or 2nd Training, in terms of measuring semantic similarities of sentence pairs. But for RoBERTa-large, according to the SVD visualization, the vector space used for 3rd Training contain entries distributed on different semantic axes in a slightly correlated manner. Moreover, unlike BERT-large-uncased, the rotated ellipse after whitening does not have the protruding patterns, and thus ICA-transformed vector space is not effective for training the encoder of RoBERTa-large. However, this is the strength of PST --- it can develop the vector space of entry embeddings naturally into the one catering for training a particular encoder, and in this case, a slightly rotated ellipse for RoBERTa-large. This pattern can be further observed in the next section involving data-augmented RoBERTa-large.

Excessive separate training of PST will diminish the capability of encoders to encode good-quality sentence embeddings. Figure \ref{fig:raw-4th} shows the vector spaces of entry embeddings built from the encoders after 3rd Training of PST for 4th Training. As for BERT encoders that rely on ICA-transformed vector spaces for training, the rationale is explained by their PCA visualizations: the protruding pattern as shown in Figure \ref{fig:svd-pca-BL} disappear. As for RoBERTa encoders that rely on quasi-isotropic vector spaces for training, deterioration is the root cause: the vector space used for RoBERTa-base degenerates into a rotated ellipse; the one for RoBERTa-large regresses into the aperture of a cone.

\begin{table*}[t]
\centering
\caption{The best results out of 10 runs are reported for $DefSent+$ encoders trained from data-augmented models: $SIMCSE$ (unsupervised) \citep{simcse}, $SNCSE$ \citep{sncse}, and $SynCSE$ \citep{syncse}. The results of data-augmented models are based on their released encoders. For $SIMCSE$ and $SynCSE$, we also attempt Prompt pooling to encode sentence embeddings, and report the results accordingly if average scores overtake their previous ones.}
\label{tab:cse-models}
\resizebox{1.0\linewidth}{!}{%
\begin{tabular}{lcccccccc}
\hline
Encoder & STS12 & STS13 & STS14 & STS15 & STS16 & STS-B & SICK-R & Avg. \\ \hline
\begin{tabular}[l]{@{}l@{}}$SIMCSE_{BERT-base}$\end{tabular}     & 68.40 & 82.41 & 74.38 & 80.91 & 78.56 & 76.85 & 72.23 & 76.25  \\ 
                                                                                    \begin{tabular}[l]{@{}l@{}}$SIMCSE_{BERT-large}$\end{tabular}    & 73.54 & 85.07 & 76.72 & 84.45 & 79.05 & 80.94 & 71.18 & 78.71  \\
                                                                                    \begin{tabular}[l]{@{}l@{}}$SIMCSE_{RoBERTa-base}$\end{tabular}  & 72.72 & 82.14 & 72.99 & 81.83 & 79.65 & 79.79 & 70.68 & 77.11  \\
                                                                                    \begin{tabular}[l]{@{}l@{}}$SIMCSE_{RoBERTa-large}$\end{tabular} & 72.86 & 83.99 & 75.62 & 84.77 & 81.80 & 81.98 & 71.26 & 78.90  \\ \hdashline
\begin{tabular}[l]{@{}l@{}}$SNCSE_{BERT-base}$\end{tabular}     & 70.67 & 84.79 & 76.99 & 83.69 & 80.51 & 81.35 & 74.77 & 78.97  \\ 
                                                                                    \begin{tabular}[l]{@{}l@{}}$SNCSE_{BERT-large}$\end{tabular}    & 71.94 & 86.66 & 78.84 & 85.74 & 80.72 & 82.29 & 75.11 & 80.19  \\
                                                                                    \begin{tabular}[l]{@{}l@{}}$SNCSE_{RoBERTa-base}$\end{tabular}  & 70.62 & 84.42 & 77.24 & 84.85 & 81.49 & 83.07 & 72.92 & 79.23  \\
                                                                                    \begin{tabular}[l]{@{}l@{}}$SNCSE_{RoBERTa-large}$\end{tabular} & 73.71 & 86.73 & 80.35 & 86.80 & 83.06 & 84.31 & 77.43 & 81.77  \\ \hdashline
\begin{tabular}[l]{@{}l@{}}$SynCSE(partial)_{RoBERTa-base}$\end{tabular}  & 77.73 & 82.10 & 78.90 & 84.53 & 82.44 & 82.49 & 80.43 & 81.23  \\
                                                                                    \begin{tabular}[l]{@{}l@{}}$SynCSE(partial)_{RoBERTa-large}$\end{tabular} & 75.95 & 84.95 & 80.20 & 85.80 & 84.52 & 85.26 & 82.20 & 82.70  \\ \hline
\begin{tabular}[l]{@{}l@{}}*$DefSent+^{1st-(CLS, AMP)-a}_{simcse-BERT-base}$\end{tabular}   & 72.59 & 85.63 & 78.05 & 84.27 & 81.05 & 82.46 & 74.28 & 79.76  \\ 
                                                                                    \begin{tabular}[l]{@{}l@{}}*$DefSent+^{1st-(CLS, AMP)-a}_{simcse-BERT-large}$\end{tabular}
                                                                                    & 74.46 & 87.40 & 79.86 & 85.20 & 82.21 & 83.23 & 75.01 & 81.05 \\
                                                                                    \begin{tabular}[l]{@{}l@{}}*$DefSent+^{1st-(CLS, AC)-a}_{simcse-RoBERTa-base}$\end{tabular}
                                                                                    & 74.33 & 85.16 & 77.60 & 84.09 & 81.97 & 81.70 & 75.14 & 80.00 \\
                                                                                    \begin{tabular}[l]{@{}l@{}}$DefSent+^{1st-(CLS, AC)-a}_{simcse-RoBERTa-large}$\end{tabular}
                                                                                    & 75.35 & 86.33 & 79.82 & 86.31 & 83.62 & 83.94 & 76.46 & 81.69 \\ \hdashline
\begin{tabular}[l]{@{}l@{}}*$DefSent+^{1st-(CLS, AMP)-a}_{sncse-BERT-base}$\end{tabular}    & 73.07 & 86.41 & 79.58 & 84.69 & 81.61 & 83.00 & 74.94 & 80.47 \\ 
                                                                                    \begin{tabular}[l]{@{}l@{}}*$DefSent+^{1st-(CLS, AMP)-a}_{sncse-BERT-large}$\end{tabular}
                                                                                    & 74.79 & 87.37 & 80.48 & 86.26 & 82.32 & 83.36 & 76.05 & 81.52 \\
                                                                                    \begin{tabular}[l]{@{}l@{}}*$DefSent+^{1st-(Mean, AC)-a}_{sncse-RoBERTa-base}$\end{tabular}
                                                                                    & 72.63 & 85.83 & 79.23 & 84.05 & 82.28 & 83.31 & 74.61 & 80.28 \\
                                                                                    \begin{tabular}[l]{@{}l@{}}$DefSent+^{1st-(CLS, AC)-a}_{sncse-RoBERTa-large}$\end{tabular}
                                                                                    & 75.34 & 87.05 & 80.93 & 85.75 & 83.11 & 84.38 & 78.30 & 82.12 \\ \hdashline
\begin{tabular}[l]{@{}l@{}}*$DefSent+^{1st-(Mean, AC)-a}_{syncse-partial-RoBERTa-base}$\end{tabular}
                                                                                    & 77.32 & 83.94 & 81.10 & 85.99 & 83.73 & 83.82 & 79.50 & 82.20 \\
                                                                                    \begin{tabular}[l]{@{}l@{}}$DefSent+^{1st-(CLS, AC)-a}_{syncse-partial-RoBERTa-large}$\end{tabular}
                                                                                    & 75.63 & 86.21 & 80.18 & 85.62 & 84.42 & 85.03 & 78.75 & 82.26 \\ \hline
\begin{tabular}[l]{@{}l@{}}*$DefSent+^{2nd-(CLS, AC)-a}_{simcse-RoBERTa-large}$\end{tabular}    & 75.97 & 87.21 & 80.76 & 86.51 & 82.63 & 84.41 & 77.97 & 82.21 \\ \hdashline
                                                                                    \begin{tabular}[l]{@{}l@{}}*$DefSent+^{2nd-(Mean, AC)-a}_{sncse-RoBERTa-large}$\end{tabular}
                                                                                    & 76.28 & 87.29 & 81.16 & 85.26 & 82.91 & 84.57 & 77.75 & 82.17 \\ \hdashline
                                                                                    \begin{tabular}[l]{@{}l@{}}*$DefSent+^{2nd-(CLS, AC)-a}_{syncse-partial-RoBERTa-large}$\end{tabular}
                                                                                    & 76.47 & 87.28 & 82.26 & 87.65 & 85.98 & 86.20 & 80.87 & 83.82 \\ \hline
\end{tabular}%
}
\end{table*}

\begin{figure*}[t!]
\centering
\includegraphics[width=0.9\linewidth]{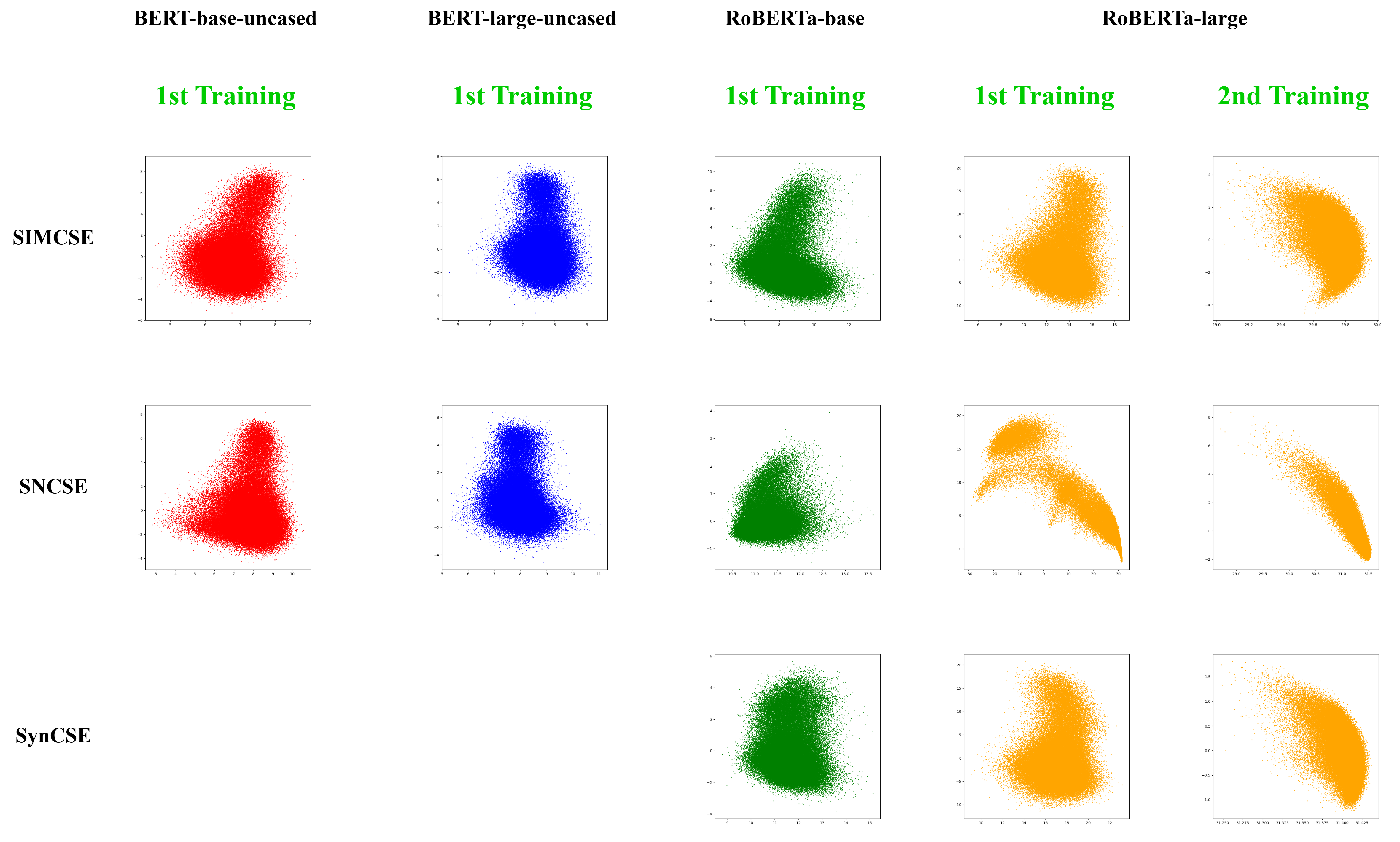}
\vspace{2em}
\caption{The vector spaces of entry embeddings used to train the encoders of the data-augmented models.}
\label{fig:CSE}
\end{figure*}

\subsection{DefSent+ for data-augmented models}
\label{sec:data-augmented-models}
In this section, we apply $DefSent+$ to data-augmented models, and the experimental results of STS tasks are reported in Table \ref{tab:cse-models}. Three sets of data-augmented models are further trained by $DefSent+$: unsupervised $SIMCSE$ \citep{simcse} (the classical one preceding other CSE series), $SNCSE$ \citep{sncse} and $SynCSE$ \citep{syncse} (two latest published ones). The entry embeddings used for training are visualized in Figure \ref{fig:CSE}.

Compared to their raw pre-trained counterparts, both difference and invariability are observed. For those previously relying on ICA-transformed (BERT encoders) and quasi-isotropic (RoBERTa-base encoder) vector spaces for training, the optimal distribution in vector spaces turn into a gourd-like shape (a big sphere-like at bottom with a small one at top). There is no sign of correlation between the distributions on different semantic axis. Plus, if the weight matrix of entry embeddings is centered first and then used for training, same level performance on STS tasks in Table \ref{tab:cse-models} can be achieved. For these two considerations, The gourd-like distribution can also be considered as quasi-isotropic. The invariable pattern is observed from the optimal vector space of entry embeddings used for training RoBERTa-large encoders. To achieve the highest performance on STS tasks, 2nd Training of PST is needed, where entry distributions resemble rotated ellipse with a sign of negative correlation. The consistency proves that RoBERTa-large encoders prefer this type of embeddings, and we will present in the next section that RoBERTa-large encoders trained using this type of embeddings are also powerful on the capability of feature-based transfer.

The performance increments as shown in Table \ref{fig:CSE} are significant. Except for $DefSent+^{2nd-(Mean, AC)-a}_{sncse-RoBERTa-large}$ that improves the previous result by 0.4\% ($82.17\% - 81.77\%$), the rest improvement ranges from 0.97\% ($82.20\% - 81.23\%$) to 3.51\% ($79.76\% - 76.25\%$). To the best of our knowledge, as of writing, in terms of $DefSent+^{2nd-(CLS, AC)-a}_{syncse-partial-RoBERTa-large}$, the average result of STS tasks (83.82\%) is the state-of-the-art performance achieved by the encoders trained without using large manually labeled datasets. There is one variation regarding encoding combinations. For further training $SNCSE_{RoBERTa-large}$, the encoding combination (Mean, AC) of 2nd Training is different from the one (CLS, AC) of 1st Training, but considering the fact that the total PST steps for further training data-augmented models are at most two, there is no big issue of consuming computational resources for one additional attempt. It seems that there is no noticeable performance difference between $DefSent+^{1st-(CLS, AC)-a}_{sncse-RoBERTa-large}$ and $DefSent+^{2nd-(Mean, AC)-a}_{sncse-RoBERTa-large}$; however, the statistical result of $DefSent+^{1st-(CLS, AC)-a}_{sncse-RoBERTa-large}$ ($81.81\pm0.45$ in terms of average) is not comparable to the one of $DefSent+^{2nd-(Mean, AC)-a}_{sncse-RoBERTa-large}$ (see Table \ref{tab:supplementary} in Appendix \ref{sec:supplementary-result}).

\begin{table*}[t]
\centering
\caption{Accuracy (\%) performance on SentEval benchmarks with the default configuration is reported. The results of previous works are based on their released encoders, and the results of our work are based on the encoders to be released. For $SIMCSE$ and $SynCSE$ encoders, we attempt Prompt pooling to encode sentence embeddings, and report the results accordingly if average scores overtake their previous ones. For $SNCSE$ encoders, we attempt CLS or Mean pooling to encode sentence embeddings, and report the results accordingly if average scores overtake their previous ones. The results of $DefSent$ are based on their best pooling strategies.}
\label{tab:senteval}
\resizebox{1.0\linewidth}{!}{%
\begin{tabular}{lcccccccc}
\hline
Encoder & MR & CR & SUBJ & MPQA & SST-2 & TREC & MRPC & Avg. \\ \hline
\begin{tabular}[l]{@{}l@{}}$DefSent_{BERT-base}$\end{tabular}    & 81.72 & 88.14 & 94.89 & 90.11 & 87.04 & 89.20 & 75.83 & 86.70  \\ 
\begin{tabular}[l]{@{}l@{}}$DefSent_{BERT-large}$\end{tabular}    & 85.46 & 90.31 & 95.43 & 90.25 & 90.39 & 90.60 & 73.74 & 88.03  \\ 
\begin{tabular}[l]{@{}l@{}}$DefSent_{RoBERTa-base}$\end{tabular}   & 84.83 & 90.86 & 94.70 & 90.68 & 90.06 & 92.60 & 76.93 & 88.67  \\ 
\begin{tabular}[l]{@{}l@{}}$DefSent_{RoBERTa-large}$\end{tabular}   & 85.99 & 90.91 & 95.28 & 91.03 & 90.12 & 92.80 & 73.28 & 88.49  \\ \hdashline
\begin{tabular}[l]{@{}l@{}}*$DefSent+^{3rd-(CLS, AMP)-a}_{BERT-base(ICA)}$\end{tabular}    & 81.06 & 87.15 & 94.59 & 89.39 & 86.55 & 89.60 & 75.94 & 86.33  \\ 
\begin{tabular}[l]{@{}l@{}}*$DefSent+^{3rd-(CLS, AMP)-a}_{BERT-large(ICA)}$\end{tabular}    & 83.43 & 89.78 & 95.26 & 90.58 & 89.18 & 89.60 & 77.68 & 87.93  \\ 
\begin{tabular}[l]{@{}l@{}}*$DefSent+^{3rd-(Mean, AC)-a}_{RoBERTa-base}$\end{tabular}   & 86.04 & 91.81 & 94.37 & 89.59 & 92.20 & 90.40 & 74.67 & 88.44  \\ 
\begin{tabular}[l]{@{}l@{}}*$DefSent+^{3rd-(Mean, AC)-a}_{RoBERTa-large}$\end{tabular}   & 87.82 & 92.05 & 94.22 & 90.66 & 93.08 & 91.80 & 77.62 & 89.61  \\ \hline
\begin{tabular}[l]{@{}l@{}}$SIMCSE_{BERT-base}$\end{tabular}     & 81.18 & 86.20 & 94.43 & 88.87 & 85.50 & 89.80 & 74.49 & 85.78  \\ 
\begin{tabular}[l]{@{}l@{}}$SIMCSE_{BERT-large}$\end{tabular}    & 85.36 & 89.43 & 95.39 & 89.68 & 90.44 & 91.80 & 76.41 & 88.36  \\ 
\begin{tabular}[l]{@{}l@{}}$SIMCSE_{RoBERTa-base}$\end{tabular}  & 83.27 & 89.54 & 93.86 & 88.51 & 88.14 & 84.80 & 75.30 & 86.20  \\ 
\begin{tabular}[l]{@{}l@{}}$SIMCSE_{RoBERTa-large}$\end{tabular} & 85.07 & 90.31 & 93.82 & 89.95 & 91.10 & 92.20 & 76.64 & 88.44  \\ \hdashline
\begin{tabular}[l]{@{}l@{}}*$DefSent+^{1st-(CLS, AMP)-a}_{simcse-BERT-base}$\end{tabular}     & 80.99 & 87.13 & 94.59 & 89.80 & 86.05 & 90.20 & 74.43 & 86.17  \\ 
\begin{tabular}[l]{@{}l@{}}*$DefSent+^{1st-(CLS, AMP)-a}_{simcse-BERT-large}$\end{tabular}    & 85.06 & 90.09 & 95.70 & 90.20 & 90.44 & 92.60 & 76.17 & 88.61  \\ 
\begin{tabular}[l]{@{}l@{}}*$DefSent+^{1st-(CLS, AC)-a}_{simcse-RoBERTa-base}$\end{tabular}  & 85.40 & 91.26 & 93.96 & 89.58 & 91.49 & 87.80 & 77.74 & 88.18  \\ 
\begin{tabular}[l]{@{}l@{}}*$DefSent+^{2nd-(CLS, AC)-a}_{simcse-RoBERTa-large}$\end{tabular} & 86.58 & 90.97 & 93.37 & 90.31 & 92.59 & 89.40 & 76.93 & 88.59  \\ \hline
\begin{tabular}[l]{@{}l@{}}$SNCSE_{BERT-base}$\end{tabular}     & 81.22 & 85.88 & 93.48 & 89.68 & 85.67 & 87.60 & 76.58 & 85.73  \\
\begin{tabular}[l]{@{}l@{}}$SNCSE_{BERT-large}$\end{tabular}    & 82.42 & 88.08 & 94.12 & 89.58 & 86.55 & 87.40 & 75.36 & 86.22  \\
\begin{tabular}[l]{@{}l@{}}$SNCSE_{RoBERTa-base}$\end{tabular}  & 83.11 & 89.11 & 92.91 & 88.93 & 86.38 & 88.00 & 76.23 & 86.38  \\
\begin{tabular}[l]{@{}l@{}}$SNCSE_{RoBERTa-large}$\end{tabular} & 85.75 & 90.49 & 93.69 & 89.89 & 90.99 & 91.80 & 74.78 & 88.20  \\ \hdashline
\begin{tabular}[l]{@{}l@{}}*$DefSent+^{1st-(CLS, AMP)-a}_{sncse-BERT-base}$\end{tabular}     & 80.95 & 86.89 & 94.56 & 89.37 & 86.11 & 88.80 & 76.41 & 86.16  \\
\begin{tabular}[l]{@{}l@{}}*$DefSent+^{1st-(CLS, AMP)-a}_{sncse-BERT-large}$\end{tabular}    & 84.95 & 89.80 & 95.33 & 89.83 & 89.57 & 92.60 & 75.54 & 88.23  \\
\begin{tabular}[l]{@{}l@{}}*$DefSent+^{1st-(Mean, AC)-a}_{sncse-RoBERTa-base}$\end{tabular}  & 85.02 & 90.99 & 93.97 & 89.44 & 90.33 & 88.80 & 77.86 & 88.06  \\
\begin{tabular}[l]{@{}l@{}}*$DefSent+^{2nd-(Mean, AC)-a}_{sncse-RoBERTa-large}$\end{tabular} & 86.28 & 91.55 & 93.73 & 90.66 & 91.76 & 92.60 & 78.20 & 89.25  \\ \hline
\begin{tabular}[l]{@{}l@{}}$SynCSE(partial)_{RoBERTa-base}$\end{tabular}  & 85.77 & 91.63 & 93.21 & 90.23 & 90.94 & 93.00 & 77.22 & 88.86  \\
\begin{tabular}[l]{@{}l@{}}$SynCSE(partial)_{RoBERTa-large}$\end{tabular} & 88.70 & 92.80 & 95.02 & 91.05 & 93.63 & 95.20 & 77.04 & 90.49  \\ \hdashline
\begin{tabular}[l]{@{}l@{}}*$DefSent+^{1st-(Mean, AC)-a}_{syncse-partial-RoBERTa-base}$\end{tabular}  & 85.70 & 91.50 & 93.82 & 90.69 & 91.32 & 92.20 & 77.80 & 89.00  \\
\begin{tabular}[l]{@{}l@{}}*$DefSent+^{2nd-(CLS, AC)-a}_{syncse-partial-RoBERTa-large}$\end{tabular} & 87.88 & 92.13 & 94.53 & 90.82 & 92.86 & 94.20 & 77.68 & 90.01  \\ \hline
\end{tabular}%
}
\end{table*}

\section{SentEval benchmarks}
\label{sec:senteval}
We evaluate the capability of feature-based transfer for 14 $DefSent+$ encoders on SentEval \citep{senteval} benchmarks. All the encoders are tested with the default configuration\footnote{\url{https://github.com/facebookresearch/SentEval}} run on the same GPU machine. Although the capability of feature-based transfer is not related to the major application of our sentence embeddings, it is still meaningful to further verify whether sentence embeddings encoded by $DefSent+$ encoders carries useful semantics, also as a way to double-confirm if data-augmented encoders are improved by $DefSent+$ or not.

The results are reported in Table \ref{tab:senteval}. As shown in the table, sentence embeddings encoded by $DefSent$ encoders are more powerful than the ones encoded by data-augmented encoders ($SynCSE(partial)_{RoBERTa-large}$ is the only exception), but are incomparable in terms of the performance on STS tasks regarding measuring similarities of sentence pairs. One the other hand, sentence embeddings encoded by $DefSent+$ encoders not only outperform $DefSent$ significantly on STS tasks, but also are comparable to $DefSent$ on the capability of feature-based transfer for various NLP downstream tasks. Four $DefSent+$ encoders achieve 89.00\%+ average results. Except $DefSent+^{2nd-(CLS, AC)-a}_{syncse-partial-RoBERTa-large}$ is outperformed by $SynCSE(partial)_{RoBERTa-large}$\footnote{When CLS pooling is used to encode sentence embeddings, $DefSent+^{2nd-(CLS, AC)-a}_{syncse-partial-RoBERTa-large}$ outperforms $SynCSE(partial)_{RoBERTa-large}$ by 0.65\% ($89.34\% - 88.69\%$)}, data-augmented encoders further trained by $DefSent+$ consistently outperform their previous counterparts in terms of average results, especially highlighted by the improvement achieved by RoBERTa-base encoders. This double-confirms that data-augmented encoders are improved by $DefSent+$. For the representative MRPC task, seven $DefSent+$ encoders achieve the results over 77.60\%, which are noticeably higher than 75.20\% (purely based on lexical overlap features, as a token of textual similarities for semantic similarities) \citep{pefbat}. 

\section{Conclusion}
\label{sec:conclusion}
In this paper, to improve sentence embeddings of language models for measuring semantic similarities of sentence pairs, we sought to excavate the potential of \citet{defsent}'s approach by projecting definition sentences into the vector space of dictionary entries. This approach belongs to self-supervised learning without using manually labeled datasets. Our exploration, $DefSent+$\footnote{Our code and encoders are publicly available under \url{https://github.com/ryuliuxiaodong/DefSent-Plus}.}, started from two perspectives: 1) full exploitation of any given dictionary dataset; 2) the vector space of entry embeddings with entry distribution approaching directionally uniform.

To accomplish the 1), we proposed Equation \ref{equ:e1} to build entry embeddings by means of sentence embeddings. As shown by the experimental results in Table \ref{tab:raw-models}, with the basic level of dictionary resource (Dataset $c$), $DefSent+$ can already encode sentence embeddings better than $DefSent$. It was validated that entry embeddings built by Equation \ref{equ:e1} are more effective for this approach than word embeddings of language models adopted by $DefSent$. Since dictionary resources can be fully exploited via this method, $DefSent+$ encoders trained with bigger datasets (Datasets $b$ and $a$) can encode sentence embeddings of better quality proportional to the size of dataset used for training.

To accomplish the 2), we proposed a novel method named PST (Progressive Separate Training) to progressively build the vector space of entry embeddings into a quasi-isotropic one. The quasi-isotropic can be further processed by the ICA algorithm more effectively for training some encoders. The strength of PST is that it can develop the vector space of entry embeddings naturally into the optimal one catering for training a particular encoder. To recapitulate, there are three types of vector space for training different encoders. First, the quasi-isotropic one resembles a ellipse or gourd for training RoBERTa-base encoder or data-augmented BERT-base-uncased, BERT-large-uncased, and RoBERTa-base encoders. Second, the developed vector space after whitening process has protruding patterns (highlighted in Figure \ref{fig:svd-pca-BL}), which can be processed by the ICA algorithm to train BERT-base-uncased and BERT-large-uncased encoders. Third, the developed vector space resembles a rotated ellipse with a sign of correlation on entry distribution, which is particularly effective for training RoBERTa-large and data-augmented RoBERTa-large encoders. $DefSent+$ encoders trained with their corresponding optimal vector spaces of entry embeddings achieve the highest average results on STS tasks shown in Tables \ref{tab:raw-models} and \ref{tab:cse-models}. Whether sentence embeddings encoded by $DefSent+$ encoders carry useful semantic information for NLP downstream tasks is verified by SentEval benchmarks presented in Table \ref{tab:senteval}.

Among the approaches without using large manually labeled datasets, $DefSent+$ encoders achieve state-of-the-art performance on both STS tasks. The performance on STS tasks is achieved by further training data-augmented RoBERTa-large encoders based on unsupervised contrastive learning.

\section{Discussion \& future directions}
\label{sec:discussion}
$DefSent+$ could be considered as a simple yet effective approach to improving sentence embeddings of language models. Based on the performance from further trained data-augmented models, $DefSent+$ provides one more promising option for other natural languages to adopt if large manually labeled datasets are not available. One can train their encoders by data augmentation methods using contrastive learning first, and then employ $DefSent+$ to perform a further optimization. Beneficially, better semantic features conveyed in sentence embeddings are attainable for measuring similarities of sentence pairs and other NLP downstream tasks.

\textbf{Data Availability.} It can be argued that digital dictionary resources might be not available in all countries. Still, dictionary resources are comparatively easy to obtain, as there is no manual effort greatly in need to make annotations or labels. Moreover, the English dictionary resources we use in this paper are commonly-used ones, which are publicly available and not particularly tailored for our use.

\textbf{Data Leakage.} We confirm from two perspectives that there is \textbf{NO} data leakage (in machine learning sense) with regard to $DefSent+$. First, it is obvious that our training data are completely independent from test data --- the training dataset comprises definition sentences while the test dataset contains sentences from task datasets (STS or SentEval). They were originally collected for different use cases. This further verifies that the training of $DefSent+$ generalizes well to unseen sentences. Second, it is widely acknowledged that neural networks entail the attribute called blackbox. When it comes to embeddings, the names of dimensionalities $\mathbb{d}$ (i.e., random variables; e.g., 768 $\mathbb{d}$ in BERT-base-uncased model) are unknown to us. Therefore, it is unviable to pre-define or select the $\mathbb{d}$ that are useful for the predictions of target labels. Furthermore, in this paper, we input entire embeddings into classifiers for various regression tasks without cherry-picking the $\mathbb{d}$.

\textbf{Training Consumption.} Our GPU devices used for training are not the most advanced ones, but still acceptable in terms of the biggest Dataset $a$ with 1-epoch gradient descent: 48 minutes (RTX 2080 Ti) for BERT-base-uncased, 49 minutes (TITAN RTX) for RoBERTa-base, and 75 minutes (GeForce RTX 3090) for BERT-large-uncased and RoBERTa-large.

\textbf{Future Works.} We will apply $DefSent+$ to other natural languages. Also, another two future directions, not the focus of this paper, can be investigated as well: exploration on the combination of other dictionary resources (as stated above, we focus on the commonly-used ones in this paper); application to reverse dictionaries (as explained in Section \ref{sec:pst}, in this paper, our focus is to encode sentence embeddings after training, and thus the optimized parameters of the pooler are no longer useful).

\appendix

\section{Hyperparameter}
\label{sec:hyperparameter}
In this paper, there are two topics involving hyperparameters.

As for the encoders such as BERT-base-uncased and BERT-large-uncased that need ICA-transformed vector spaces to train with for optimal performance, same as \citet{ICA-transform}, we use FastICA in Scikit-learn \citep{Scikit-learn}. Almost every setting is default except for two customized ones. First, as suggested in \citet{ICA-transform}'s paper, we set the maximum iteration as 1000 for transformation, and 10000 or 200 is not effective for our training objective. Second, after transforming weight matrix of entry embeddings is finished, the matrix should be scaled up by multiplying 100, because typically after ICA-transformation, the dimensional values are approximately 1\% of original scale --- a clear official example can be found under \href{https://scikit-learn.org/stable/auto_examples/decomposition/plot_ica_blind_source_separation.html}{https://scikit-learn.org/stable/auto\_examples/decomposition/plot\_ica\_blind\_source\_separation.html}. For reproducibility, we set the random state as commonly-used 42, and CPU is Intel(R) Core(TM) i9-9980XE CPU \ead{@} 3.00GHz.

As for training different encoders, we follow the default optimizer setting of fine-tuning BERT \citep{bert}. Two settings are flexible. First, the batch size is 32 for Dataset $a$ and 16 for Datasets $b$ and $c$. Second, the learning rates are summarized in Table \ref{tab:learning-rate}. There is a pattern for learning rates compared to the ones used in previous steps --- the learning rate used at the current step of training should be no bigger than the previous one. For training raw pre-trained models, the learning rates used for 1st Training are determined through grid search within \{5e-5, 4e-5, 3e-5\}; for data-augmented models, we take a reference to their original papers, and make slight modifications if necessary.

There is only 1-epoch gradient descent required for this approach. For all encoders to be trained, regardless of model type or dataset size, the performance on STS tasks gradually decreases as epoch number grows. From our understanding, the training objective of this approach is slightly different from the MLM pre-training objective of auto-encoding models. This is not merely because the prediction layer is frozen for this approach. For the later, the error loss should be the smaller the better with substantial pre-training iterations. For this approach, there is a special consideration. For example, the sentence embedding of ``a long-necked African animal'' can be considered as good after its projection into the vector space of entry embeddings via the pooler, if the projected representation generates evenly-distributed softmax scores for the entry embeddings of ``giraffe'', ``vulture'', and ``ostrich'', as well as relatively small scores to other entries like ``lion'', ``hyena'', and so on. To prevent the softmax score from being biased towards a particular entry, 1-epoch gradient descent is a practical solution.

\begin{table*}[t]
\centering
\caption{Learning rates for encoders trained at different steps of PST.}
\label{tab:learning-rate}
\resizebox{0.6\linewidth}{!}{%
\begin{tabular}{lccc}
\hline
Encoder & 1st Training & 2nd Training & 3rd Training \\ \hline
\begin{tabular}[l]{@{}l@{}}$DefSent+_{BERT-base}$\end{tabular}    & 5e-5 & 4e-5 & 3e-5  \\ 
\begin{tabular}[l]{@{}l@{}}$DefSent+_{BERT-large}$\end{tabular}    & 4e-5 & 4e-5 & 3e-5  \\ 
\begin{tabular}[l]{@{}l@{}}$DefSent+_{RoBERTa-base}$\end{tabular}   & 3e-5 & 2e-5 & 1e-5  \\ 
\begin{tabular}[l]{@{}l@{}}$DefSent+_{RoBERTa-large}$\end{tabular}   & 3e-5 & 2e-5 & 1e-5  \\ \hdashline
\begin{tabular}[l]{@{}l@{}}$DefSent+_{simcse-BERT-base}$\end{tabular}     & 3e-5 & - & -  \\ 
\begin{tabular}[l]{@{}l@{}}$DefSent+_{simcse-BERT-large}$\end{tabular}    & 1e-5 & - & -  \\ 
\begin{tabular}[l]{@{}l@{}}$DefSent+_{simcse-RoBERTa-base}$\end{tabular}  & 1e-5 & - & -  \\ 
\begin{tabular}[l]{@{}l@{}}$DefSent+_{simcse-RoBERTa-large}$\end{tabular} & 3e-5 & 2e-5 & -  \\ \hdashline
\begin{tabular}[l]{@{}l@{}}$DefSent+_{sncse-BERT-base}$\end{tabular}     & 2e-5 & - & -  \\
\begin{tabular}[l]{@{}l@{}}$DefSent+_{sncse-BERT-large}$\end{tabular}    & 1e-5 & - & -  \\
\begin{tabular}[l]{@{}l@{}}$DefSent+_{sncse-RoBERTa-base}$\end{tabular}  & 2e-6 & - & -  \\
\begin{tabular}[l]{@{}l@{}}$DefSent+_{sncse-RoBERTa-large}$\end{tabular} & 1e-6 & 1e-6 & -  \\ \hdashline
\begin{tabular}[l]{@{}l@{}}$DefSent+_{syncse-RoBERTa-base}$\end{tabular}  & 1e-6 & - & -  \\
\begin{tabular}[l]{@{}l@{}}$DefSent+_{syncse-RoBERTa-large}$\end{tabular} & 2e-5 & 5e-6 & -  \\ \hline
\end{tabular}%
}
\end{table*}

\begin{table*}[t]
\centering
\caption{Supplementary results of STS tasks for $DefSent+$ encoders to be released.}
\label{tab:supplementary}
\resizebox{1.0\linewidth}{!}{%
\begin{tabular}{lcccccccc}
\hline
Encoder & STS12 & STS13 & STS14 & STS15 & STS16 & STS-B & SICK-R & Avg. \\ \hline
\begin{tabular}[l]{@{}l@{}}*$DefSent+^{3rd-(CLS, AMP)-a}_{BERT-base(ICA)}$\end{tabular}    & 71.32 & 84.67 & 77.02 & 83.53 & 81.15 & 82.12 & 73.07 & 78.98  \\ 
\begin{tabular}[l]{@{}l@{}}*$DefSent+^{3rd-(CLS, AMP)-a}_{BERT-large(ICA)}$\end{tabular}    & 72.83 & 86.92 & 79.33 & 84.52 & 81.68 & 82.28 & 74.94 & 80.36  \\ 
\begin{tabular}[l]{@{}l@{}}*$DefSent+^{3rd-(Mean, AC)-a}_{RoBERTa-base}$\end{tabular}   & 72.67 & 84.95 & 77.04 & 83.53 & 81.16 & 81.15 & 74.50 & 79.29  \\ 
\begin{tabular}[l]{@{}l@{}}*$DefSent+^{3rd-(Mean, AC)-a}_{RoBERTa-large}$\end{tabular}   & 70.38 & 85.69 & 79.24 & 85.19 & 82.77 & 83.06 & 77.04 & 80.48  \\ \hline
\begin{tabular}[l]{@{}l@{}}*$DefSent+^{1st-(CLS, AMP)-a}_{simcse-BERT-base}$\end{tabular}     & 71.70\scriptsize$\pm$0.23 & 85.30\scriptsize$\pm$0.13 & 77.83\scriptsize$\pm$0.20 & 83.96\scriptsize$\pm$0.16 & 80.69\scriptsize$\pm$0.29 & 82.11\scriptsize$\pm$0.29 & 73.97\scriptsize$\pm$0.31 & 79.36\scriptsize$\pm$0.14  \\ 
\begin{tabular}[l]{@{}l@{}}*$DefSent+^{1st-(CLS, AMP)-a}_{simcse-BERT-large}$\end{tabular}    & 74.22\scriptsize$\pm$0.22 & 87.14\scriptsize$\pm$0.23 & 79.57\scriptsize$\pm$0.14 & 85.13\scriptsize$\pm$0.16 & 82.19\scriptsize$\pm$0.07 & 83.11\scriptsize$\pm$0.13 & 74.86\scriptsize$\pm$0.20 & 80.89\scriptsize$\pm$0.11  \\ 
\begin{tabular}[l]{@{}l@{}}*$DefSent+^{1st-(CLS, AC)-a}_{simcse-RoBERTa-base}$\end{tabular}  & 73.99\scriptsize$\pm$0.30 & 84.78\scriptsize$\pm$0.19 & 77.10\scriptsize$\pm$0.27 & 83.60\scriptsize$\pm$0.24 & 81.73\scriptsize$\pm$0.26 & 81.46\scriptsize$\pm$0.15 & 75.20\scriptsize$\pm$0.19 & 79.69\scriptsize$\pm$0.15  \\ 
\begin{tabular}[l]{@{}l@{}}*$DefSent+^{2nd-(CLS, AC)-a}_{simcse-RoBERTa-large}$\end{tabular} & 75.81\scriptsize$\pm$0.50 & 87.00\scriptsize$\pm$0.14 & 80.15\scriptsize$\pm$0.27 & 86.17\scriptsize$\pm$0.23 & 82.96\scriptsize$\pm$0.39 & 84.50\scriptsize$\pm$0.25 & 77.65\scriptsize$\pm$0.26 & 82.04\scriptsize$\pm$0.16  \\ \hdashline
\begin{tabular}[l]{@{}l@{}}*$DefSent+^{1st-(CLS, AMP)-a}_{sncse-BERT-base}$\end{tabular}     & 72.25\scriptsize$\pm$0.48 & 86.25\scriptsize$\pm$0.18 & 79.01\scriptsize$\pm$0.25 & 84.35\scriptsize$\pm$0.29 & 81.36\scriptsize$\pm$0.20 & 82.54\scriptsize$\pm$0.27 & 74.71\scriptsize$\pm$0.29 & 80.06\scriptsize$\pm$0.20  \\
\begin{tabular}[l]{@{}l@{}}*$DefSent+^{1st-(CLS, AMP)-a}_{sncse-BERT-large}$\end{tabular}    & 74.30\scriptsize$\pm$0.43 & 87.50\scriptsize$\pm$0.15 & 80.35\scriptsize$\pm$0.14 & 85.88\scriptsize$\pm$0.23 & 81.99\scriptsize$\pm$0.21 & 82.95\scriptsize$\pm$0.22 & 75.86\scriptsize$\pm$0.22 & 81.26\scriptsize$\pm$0.12  \\
\begin{tabular}[l]{@{}l@{}}*$DefSent+^{1st-(Mean, AC)-a}_{sncse-RoBERTa-base}$\end{tabular}  & 72.50\scriptsize$\pm$0.13 & 85.77\scriptsize$\pm$0.08 & 79.07\scriptsize$\pm$0.13 & 83.78\scriptsize$\pm$0.14 & 82.16\scriptsize$\pm$0.09 & 83.01\scriptsize$\pm$0.15 & 74.42\scriptsize$\pm$0.11 & 80.10\scriptsize$\pm$0.10  \\
\begin{tabular}[l]{@{}l@{}}*$DefSent+^{2nd-(Mean, AC)-a}_{sncse-RoBERTa-large}$\end{tabular} & 76.28\scriptsize$\pm$0.12 & 87.05\scriptsize$\pm$0.09 & 80.94\scriptsize$\pm$0.09 & 84.92\scriptsize$\pm$0.05 & 82.88\scriptsize$\pm$0.08 & 84.39\scriptsize$\pm$0.11 & 77.59\scriptsize$\pm$0.07 & 82.01\scriptsize$\pm$0.05  \\ \hdashline
\begin{tabular}[l]{@{}l@{}}*$DefSent+^{1st-(Mean, AC)-a}_{syncse-partial-RoBERTa-base}$\end{tabular}  & 77.23\scriptsize$\pm$0.06 & 83.76\scriptsize$\pm$0.15 & 81.08\scriptsize$\pm$0.07 & 85.99\scriptsize$\pm$0.04 & 83.64\scriptsize$\pm$0.07 & 83.75\scriptsize$\pm$0.06 & 79.49\scriptsize$\pm$0.06 & 82.13\scriptsize$\pm$0.05  \\
\begin{tabular}[l]{@{}l@{}}*$DefSent+^{2nd-(CLS, AC)-a}_{syncse-partial-RoBERTa-large}$\end{tabular} & 75.92\scriptsize$\pm$0.26 & 86.98\scriptsize$\pm$0.26 & 82.16\scriptsize$\pm$0.18 & 87.52\scriptsize$\pm$0.14 & 86.01\scriptsize$\pm$0.10 & 86.56\scriptsize$\pm$0.16 & 80.83\scriptsize$\pm$0.10 & 83.73\scriptsize$\pm$0.12  \\ \hline
\end{tabular}%
}
\end{table*}

\begin{table*}[t!]
\centering
\caption{The pooling strategies used to encode sentence embeddings for tasks are summarized in this table.}
\label{tab:pooling-summary}
\resizebox{0.44\linewidth}{!}{%
\begin{tabular}{lcc}
\hline
Encoder & STS & SentEval \\ \hline
\begin{tabular}[l]{@{}l@{}}*$DefSent+^{3rd-(CLS, AMP)-a}_{BERT-base(ICA)}$\end{tabular}    & Prompt & Mean  \\ 
\begin{tabular}[l]{@{}l@{}}*$DefSent+^{3rd-(CLS, AMP)-a}_{BERT-large(ICA)}$\end{tabular}    & Prompt & CLS  \\ 
\begin{tabular}[l]{@{}l@{}}*$DefSent+^{3rd-(Mean, AC)-a}_{RoBERTa-base}$\end{tabular}   & Prompt & Prompt  \\ 
\begin{tabular}[l]{@{}l@{}}*$DefSent+^{3rd-(Mean, AC)-a}_{RoBERTa-large}$\end{tabular}   & Mean & Prompt  \\ \hdashline
\begin{tabular}[l]{@{}l@{}}*$DefSent+^{1st-(CLS, AMP)-a}_{simcse-BERT-base}$\end{tabular}     & Prompt & CLS  \\ 
\begin{tabular}[l]{@{}l@{}}*$DefSent+^{1st-(CLS, AMP)-a}_{simcse-BERT-large}$\end{tabular}    & Prompt & CLS  \\ 
\begin{tabular}[l]{@{}l@{}}*$DefSent+^{1st-(CLS, AC)-a}_{simcse-RoBERTa-base}$\end{tabular}  & Prompt & Prompt  \\ 
\begin{tabular}[l]{@{}l@{}}*$DefSent+^{2nd-(CLS, AC)-a}_{simcse-RoBERTa-large}$\end{tabular} & Prompt & Prompt  \\ \hdashline
\begin{tabular}[l]{@{}l@{}}*$DefSent+^{1st-(CLS, AMP)-a}_{sncse-BERT-base}$\end{tabular}     & Prompt & Mean  \\
\begin{tabular}[l]{@{}l@{}}*$DefSent+^{1st-(CLS, AMP)-a}_{sncse-BERT-large}$\end{tabular}    & Prompt & CLS  \\
\begin{tabular}[l]{@{}l@{}}*$DefSent+^{1st-(Mean, AC)-a}_{sncse-RoBERTa-base}$\end{tabular}  & Prompt & Prompt  \\
\begin{tabular}[l]{@{}l@{}}*$DefSent+^{2nd-(Mean, AC)-a}_{sncse-RoBERTa-large}$\end{tabular} & Prompt & Prompt  \\ \hdashline
\begin{tabular}[l]{@{}l@{}}*$DefSent+^{1st-(Mean, AC)-a}_{syncse-partial-RoBERTa-base}$\end{tabular}  & CLS & Prompt  \\
\begin{tabular}[l]{@{}l@{}}*$DefSent+^{2nd-(CLS, AC)-a}_{syncse-partial-RoBERTa-large}$\end{tabular} & CLS & Prompt  \\ \hline
\end{tabular}%
}
\end{table*}

\begin{table*}[t]
\centering
\caption{Statistical average results of grid search on STS tasks to determine encoding combinations after 1st Training of PST based on Dataset $a$. The \textbf{bold figures} indicate optimal ones.}
\label{tab:grid-search}
\resizebox{0.8\linewidth}{!}{%
\begin{tabular}{lcccc}
\hline
Encoder & (Mean, AMP) & (CLS, AMP) & (Mean, AC) & (CLS, AC) \\ \hline
\begin{tabular}[l]{@{}l@{}}$DefSent+^{1st-unknown-a}_{BERT-base}$\end{tabular}    & 77.50\scriptsize$\pm$0.21 & \textbf{77.88}\scriptsize$\pm$0.22 & 74.62\scriptsize$\pm$0.25 & 75.10\scriptsize$\pm$0.23  \\ 
\begin{tabular}[l]{@{}l@{}}$DefSent+^{1st-unknown-a}_{BERT-large}$\end{tabular}    & 78.16\scriptsize$\pm$0.70 & \textbf{79.40}\scriptsize$\pm$0.72 & 75.27\scriptsize$\pm$0.83 & 76.54\scriptsize$\pm$0.63  \\ 
\begin{tabular}[l]{@{}l@{}}$DefSent+^{1st-unknown-a}_{RoBERTa-base}$\end{tabular}   & 76.68\scriptsize$\pm$0.29 & 76.78\scriptsize$\pm$0.22 & \textbf{78.44}\scriptsize$\pm$0.14 & 77.66\scriptsize$\pm$0.35  \\ 
\begin{tabular}[l]{@{}l@{}}$DefSent+^{1st-unknown-a}_{RoBERTa-large}$\end{tabular}   & 76.97\scriptsize$\pm$0.25 & 79.39\scriptsize$\pm$0.28 & \textbf{79.81}\scriptsize$\pm$0.19 & 79.49\scriptsize$\pm$0.26  \\ \hline
\end{tabular}%
}
\end{table*}

\section{Supplementary information}
\label{sec:supplementary-result}
Three pieces of information are supplemented in this Section. Table \ref{tab:supplementary} shows supplementary results of STS tasks for $DefSent+$ encoders to be released: for raw pre-trained models, corresponding statistical results are provided in Table \ref{tab:raw-models}; for data-augmented models, corresponding best results are provided in Table \ref{tab:cse-models}. The pooling strategies used to encode sentence embeddings for different tasks are listed in Table \ref{tab:pooling-summary}. For encoders trained from raw pre-trained models based on Dataset $c$ after 1st Training of PST, the pooling strategies are different: Mean for BERT encoders and CLS for RoBERTa encoders. The results of grid search to determine encoding combinations after 1st Training of PST are exemplified by raw pre-trained models with Dataset $a$, which is presented in Table \ref{tab:grid-search}. The search is fully conducted after the optimal learning rate for 1st Training is determined (see Appendix \ref{sec:hyperparameter}) based on (Mean, AMP).

\printcredits



\bibliographystyle{cas-model2-names}

\bibliography{cas-refs}


\end{document}